\documentclass{article} 
\usepackage{arxiv}
\setcitestyle{authoryear,open={(},close={)}}

\usepackage{fullpage}
\usepackage{amsmath,amsfonts,bm}

\def\eqref#1{equation~\ref{#1}}
\def\1{\bm{1}}

\DeclareMathAlphabet{\mathsfit}{\encodingdefault}{\sfdefault}{m}{sl}
\SetMathAlphabet{\mathsfit}{bold}{\encodingdefault}{\sfdefault}{bx}{n}

\usepackage{url}
\usepackage{graphicx}
\usepackage{wrapfig}
\usepackage{bbm}
\usepackage{amsmath}
\usepackage{amssymb}
\usepackage{mathtools}
\usepackage{amsthm}
\usepackage{enumitem}
\usepackage{natbib}
\usepackage{xcolor}
\usepackage{multirow}
\usepackage{setspace}
\usepackage{enumitem}
\usepackage{color, colortbl}
\usepackage{mathtools}
\usepackage{cleveref}

\theoremstyle{plain}

\usepackage{hyperref}
\usepackage{listings}
\usepackage{url}
\usepackage{color,graphicx}
\usepackage{siunitx}
\usepackage{soul}
\usepackage{graphics} 
\usepackage{bookmark}
\usepackage[dvipsnames, svgnames, x11names]{xcolor}
\usepackage{hyperref} 
\usepackage{textcomp}
\usepackage{multirow}
\usepackage{nth}
\usepackage{indentfirst}
\usepackage{siunitx}
\usepackage{changepage}
\usepackage{bm}
\usepackage{setspace}
\usepackage{natbib}
\usepackage{multirow}
\usepackage{multicol}
\usepackage{colortbl}
\usepackage{booktabs}
\usepackage{amsmath}
\usepackage{xcolor}
\usepackage{amsthm}
\theoremstyle{remark}

\usepackage{graphicx}
\usepackage{subfigure}
\usepackage{wrapfig}
\usepackage{enumitem}

\usepackage[most]{tcolorbox}

\definecolor{violetBg}{RGB}{245,241,250}
\definecolor{violetBorder}{RGB}{136,117,201}
\definecolor{violetAccent}{RGB}{106,90,205}
\definecolor{lightblue}{rgb}{0.933,0.968,0.988}

\usepackage[utf8]{inputenc}
\usepackage[T1]{fontenc}   
\usepackage{hyperref}      
\usepackage{url}         
\usepackage{booktabs}     
\usepackage{amsfonts}    
\usepackage{nicefrac}     
\usepackage{microtype}     
\usepackage{xcolor}       
\usepackage[table]{xcolor}
\usepackage{graphicx}
\usepackage{pifont}

\usepackage{amsmath}
\usepackage{algorithm}
\usepackage{enumerate}
\usepackage{multirow}
\usepackage{amssymb}
\usepackage[noend]{algpseudocode}

\usepackage{verbatim}

\definecolor{lightblue}{rgb}{0.933,0.968,0.988}
\definecolor{lightred}{RGB}{255,218,225}
\definecolor{lightgray}{RGB}{235,235,235}

\newcommand{\cmark}{\textcolor{green!70!black}{\ding{51}}}
\newcommand{\xmark}{\textcolor{red!80!black}{\ding{55}}}

\newtcolorbox{questionbox}[1][]{%
enhanced,
breakable,
colback=violetBg,
colframe=violetBorder,
boxrule=0.6pt,
arc=4pt,
left=8pt,right=8pt,top=8pt,bottom=8pt,
halign=center,
#1
}

\usepackage{geometry}
\usepackage{titling} 
\usepackage{setspace}

\title{\textbf{Edit-R2: Context-Aware Reinforcement Learning for Multi-Turn Image Editing}}

\vspace{-20pt}
\author{
{{\textbf{Yuxiao Ye$^{1,2}$ ~ Haoran He$^{1}$ ~ Fangyuan Kong$^{2}$ ~ Xintao Wang$^{2}$ ~ Pengfei Wan$^{2}$}}}\\
{\textbf{Kun Gai$^{2}$ ~ Ling Pan$^{1}$}} \\
{\normalsize{$^{1}$Hong Kong University of Science and Technology $^{2}$Kuaishou Technology}}
}
\date{}

\makeatletter
\renewenvironment{abstract}{
\begin{center}{\Large\bfseries Abstract}\end{center}
\vspace{-.8em}
\begingroup
\setlength{\parskip}{0.35em}
\setstretch{1.}
\normalsize
}{\par\endgroup}
\makeatother

\begin{document}

\newgeometry{left=1in,right=1in,top=0.6in,bottom=0.6in}

\maketitle

\vspace{-.5in}
\begin{abstract}
Text-guided image editing has advanced rapidly with diffusion models and unified multimodal foundation models. However, most existing methods remain confined to single-turn settings, overlooking the more realistic scenario of multi-turn in-context editing, where users iteratively refine an image through a sequence of instructions. In this setting, a model must follow each new instruction while preserving accumulated session-level constraints, challenged by two coupled failure modes: long-context dilution, where sparse textual constraints become difficult to recover from growing interleaved image-text histories, and state contamination, where earlier editing mistakes degrade subsequent generations. We introduce Edit-R2, a novel reinforcement learning post-training framework for unified multimodal models. Edit-R2 reconstructs the operative session intent, which effectively consolidates scattered historical constraints into an explicit reasoning trace before each editing turn. It further enables multi-turn RL over both reasoning and generation through a unified objective that jointly optimizes intent reconstruction generation in discrete text space and flow-matching image generation in continuous latent space, while a trajectory filtering mechanism suppresses corrupted rollouts to stabilize training under state contamination. To support systematic evaluation, we introduce MICE-Bench, a large-scale benchmark for multi-turn in-context editing with automated metrics for instruction following (IF), content consistency (CC), and global awareness (GA) over accumulated session constraints. Experiments show that Edit-R2 substantially improves multi-turn in-context editing and achieves competitive performance compared against strong baselines.

\noindent \textbf{Github:} \url{https://github.com/yuxiaooye/Edit-R2}

\begin{figure}[H]
    \centering
    \includegraphics[width=1\linewidth]{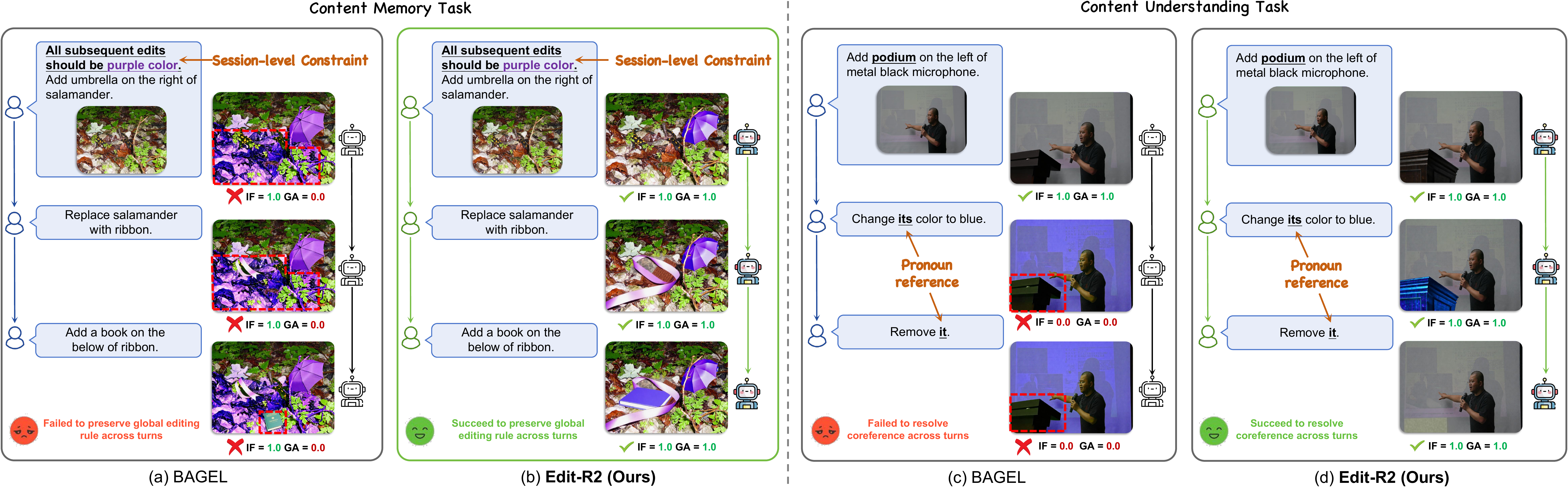}
    \vspace{-10pt}
    \caption{Overview of in-context editing. (a)-(b): \textit{Content memory} task. A session-level constraint requires all added or modified elements to be purple. BAGEL violates this constraint throughout the session, while Edit-R2 faithfully adheres to it, achieving GA\,=\,1.0 at every turn. (c)-(d): \textit{Content understanding} task. The user intends to use ``its'' to refer to the podium added in the preceding turn 1. BAGEL fails to resolve the coreference, whereas Edit-R2 correctly edits the intended object at every turn. Instruction is simplified for demonstration.}
    \label{fig:ts}
\end{figure}
\end{abstract}

\clearpage
\restoregeometry

\section{Introduction}
Text-guided image editing has advanced rapidly with diffusion and unified multi-modal foundation models~\citep{wu2026visualgenerationnewera}, which enables diverse natural-language edits such as object insertion, attribute manipulation, and semantic replacement. 
Despite its progress, the dominant formulation remains largely \emph{single-turn}: given input image(s) and one instruction, the model is expected to produce one edited image~\citep{wu2025omnigen2, icedit, yu2025anyedit, zhao2024ultraedit}. This isolated editing formulation does not reflect realistic editing workflows and leaves a key aspect of interactive editing underexplored. In practice, editing is often an iterative session: a user may first change an object, then adjust the background, and finally revise a previous preference, 
while expecting the system to preserve some of the earlier requirements and interpret new instructions without context misunderstanding. 
Therefore, a capable editing system should operate not only on the current image and instruction, but also on the accumulated session context for supporting session-level behavior and optimization.

We term this paradigm \emph{multi-turn in-context image editing} (or \textbf{\emph{in-context editing}} for brevity\footnote{Prior works~\citep{realign, wu2025omnigen2} use ``in-context'' to refer to single-turn editing conditioned on multiple reference images, which differs from our usage.}): at each turn, the model receives an interleaved history of previous instructions and generated images, and must produce an edited image that follows the new instruction while maintaining consistency with constraints introduced in prior turns (see Figure~\ref{fig:ts}). In multi-turn sessions, the current instruction may contain pronouns, ellipses, reversions, or preferences whose meaning depends on earlier turns. Meanwhile, some historical requirements should persist, whereas others may be overridden by later instructions. The current image alone may not reveal which previous constraints should remain active. 
Thus, the model must infer the operative session intent from a 
sparse, implicit, and dynamically updated image-text history.
Despite its practical importance, this in-context capability remains underexplored in current image editing models, datasets, and training paradigms. 
Most benchmarks evaluate isolated input-instruction-output triples~\citep{step1x, hui2024hq}, and existing multi-turn benchmarks~\citep{magicbrush, imgedit, pico_banana, seed-data-edit} remain \textit{history-agnostic}: each turn can be solved from the immediately preceding image and the current instruction, without requiring explicit retention of earlier constraints.

\begin{figure}[tb]
    \centering
    \includegraphics[width=1.\linewidth]{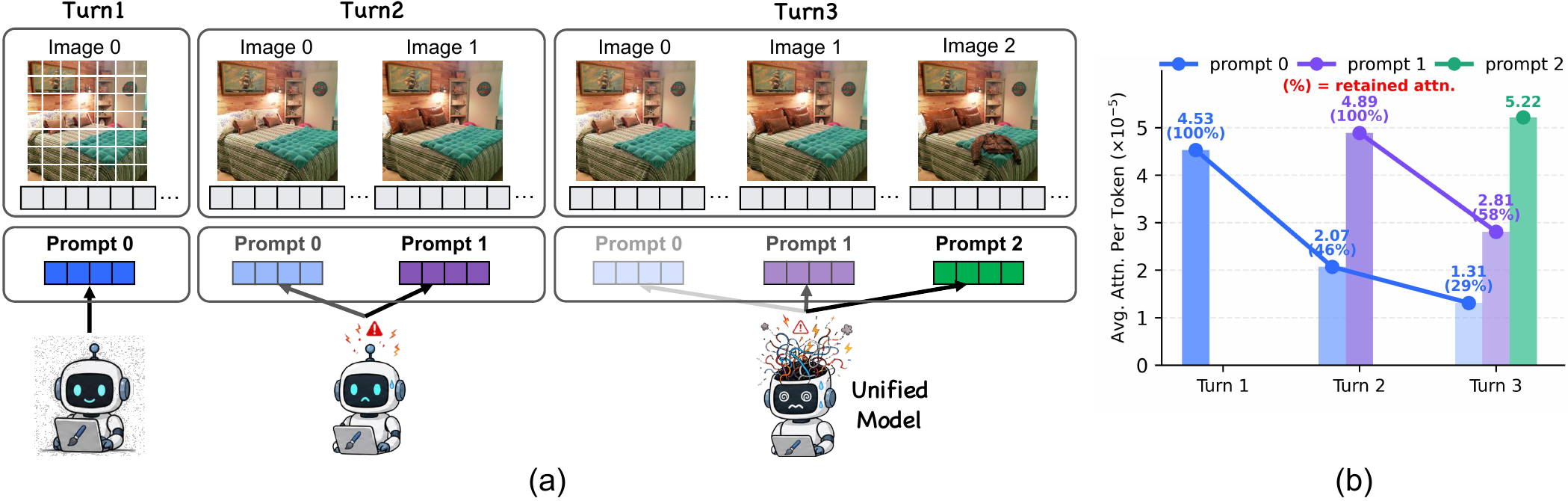}
    \vspace{-10pt}
    \caption{(a): Long-context issue for in-context editing. (b): Attention magnitude of the flow-matching generator across prompts: As editing progresses, attention to prompts from earlier turns weakens significantly, leading to context loss. 
    }
    \label{fig:long-context}
\end{figure}

Given the sequential nature of in-context editing, we formulate this as a session-level reinforcement learning (RL) problem and develop \textbf{Edit-R2}, a new RL paradigm that optimizes editing beyond traditional isolated local edits. Under this crucial scenario, each generated image becomes part of the state for future turns, and the value of an edit depends not only on its compliance with the current instruction, but also on its ability to preserve the context needed in later turns. This long-horizon dependency makes RL a natural fit. Although recent advances in RL have exhibited strong gains, they are largely limited to single-turn visual generation~\citep{flow_grpo, dancegrpo, diffusion_nft}, where the policy is optimized against a static conditioning context~\citep{editscore, uniworld_v2, vision_cot}.
Transitioning to this multi-turn paradigm poses a significant challenge, where \emph{image editing models easily get lost in the long interleaved history, leading to severe long-context intent dilution} (see Figure~\ref{fig:long-context}). 
As the session progresses, the interleaved image-text context grows rapidly, which poses critical challenges for the policy to recover the active session intent.
In unified models, this imbalance is further amplified by the heavy visual-token budget required to encode generated images~\citep{vit}, 
making critical textual constraints increasingly sparse and distant in the context across history.
Furthermore, even when an edit appears acceptable for the current instruction, any missed or incorrect detail is carried into the next turn, where it can \emph{mislead subsequent edits} and make the rollout unreliable for training.

Edit-R2 is a novel RL framework that addresses these critical bottlenecks with explicit session-intent recovery and trajectory-level stabilization. 
Built upon the unified model BAGEL~\citep{bagel}, Edit-R2 first reconstructs the operative intent for the current turn via in-context chain-of-thought (IC-CoT), which reasons over the dialogue history and consolidates scattered historical constraints into a compact reasoning trace, mitigating the dilution of early textual constraints and providing the generator with a stable state representation.
Crucially, IC-CoT cannot be optimized in isolation, as an apparently reasonable intent reconstruction is useful only if it leads to successful downstream image editing. 
To achieve the synergy, we jointly optimize IC-CoT generation and visual generation with unified multi-turn RL, which couples the discrete text space of session-intent reconstruction with the continuous latent space of the flow-matching image generator. 
To prevent corrupted rollouts from dominating training and contaminating later training steps, Edit-R2 further introduces a trajectory filtering mechanism for stable optimization.
Finally, to provide the both the training environment and evaluation protocol required for multi-turn RL and resolve the evaluation bottleneck, we construct MICE-Bench, a large-scale benchmark for in-context editing covering \emph{content memory} and \emph{content understanding} tasks with a fully automated evaluation pipeline built on comprehensive metrics: instruction following (IF), content consistency (CC)~\citep{edival}, and newly introduced global awareness (GA) measuring compliance with accumulated session-level constraints. 
Experiments on MICE-Bench demonstrate that Edit-R2 substantially outperforms existing open-source models and achieves competitive performance against closed-source counterparts.

Our contributions are summarized as follows: \textbf{(\romannumeral1)} We formulate multi-turn in-context image editing as a session-level RL problem and introduce {MICE-Bench}, the first large-scale automated benchmark for this setting, accompanied by a new GA metric that explicitly quantifies session-level constraint compliance. 
\textbf{(\romannumeral2)} We propose {Edit-R2}, a novel RL post-training framework that reconstructs the active session intent with IC-CoT and jointly optimizes intent reconstruction and visual generation across discrete text and continuous latent spaces, with trajectory filtering for stable multi-turn training. 
\textbf{(\romannumeral3)} Extensive experiments show that Edit-R2 substantially improves BAGEL \textbf{(+18\% IF, +18\% GA)} and \textbf{(+8\% IF, +15\% GA)}, achieving competitive performance over strong models.

\section{Related Work}

\paragraph{Instruction-based Image Editing Models.} 
InstructPix2Pix~\citep{brooks2023instructpix2pix} pioneered the learning-based image editing, subsequent models~\cite{sheynin2024emu, step1x} such as Qwen-Image-Edit~\citep{qwen_image_edit} and FLUX.1 Kontext~\citep{flux_kontext} extend with stronger architectures and larger-scale training.
Recently, image editing has evolved rapidly from single-turn~\cite{wu2025omnigen2, icedit} to multi-turn paradigms. 
Several training-free methods aim to mitigate quality degradation across successive turns~\citep{freqedit, sheynin2024emu}. Auto-regressive models including Nano-Banana-2~\cite{banana}, GPT-Image-1~\citep{gpt-image}, OmniGen~\citep{omnigen}, and BAGEL~\citep{bagel} enable in-context editing. Despite these advances, existing models struggle to resolve the long-horizon challenges inherent to in-context editing.

\paragraph{Image Editing Benchmarks.}
Existing benchmarks can be categorized by the degree of temporal dependency they encode (summarized in Table~\ref{tab:test_set}).
\textbf{(\romannumeral1) Single-turn.} Most early benchmarks target isolated editing instructions~\citep{hui2024hq, imgedit, step1x}.
\textbf{(\romannumeral2) Multi-turn.} MagicBrush~\citep{magicbrush}, Seed-Data-Edit~\citep{seed-data-edit} provide multi-turn corpora~\citep{pico_banana}, while MSE-Bench~\citep{vincie} and EdiVal-Bench~\citep{edival} offer evaluations. However, all of them treat each turn as independent task and requires no retention of session-level context. 
\textbf{(\romannumeral3) Multi-turn in-context.} The need for in-context image-editing evaluation has recently been recognized. ~\citep{imgedit, zhang2026non} establish the long-range dependencies across turns with several human-crafted examples. However, neither work provides a systematic and scalable evaluation protocol.

\paragraph{Reinforcement Learning for Generative Models.}
Flow-GRPO~\citep{flow_grpo} and Dance-GRPO~\citep{dancegrpo} adopt group relative policy optimization by reformulating the sampling of diffusion and rectified flows via Stochastic Differential Equations (SDEs), and Diffusion-NFT~\citep{diffusion_nft} applies direct preference optimization. In image editing, EditScore~\citep{editscore}, Edit-R1~\citep{uniworld_v2} employ RL in single-turn settings~\citep{vision_cot}.
Works such as~\citep{liu2026unigrpo, kou2026think} jointly optimize reasoning module and image generator through RL, but remain restricted to single-turn tasks.
All of the methods are history-agnostic. In contrast, Edit-R2 is one of the first RL post-training framework for in-context editing.

\section{MICE-Bench: Multi-Turn In-Context Editing Benchmark}
\label{sec:data}

\begin{figure}
    \centering
    \includegraphics[width=0.9\linewidth]{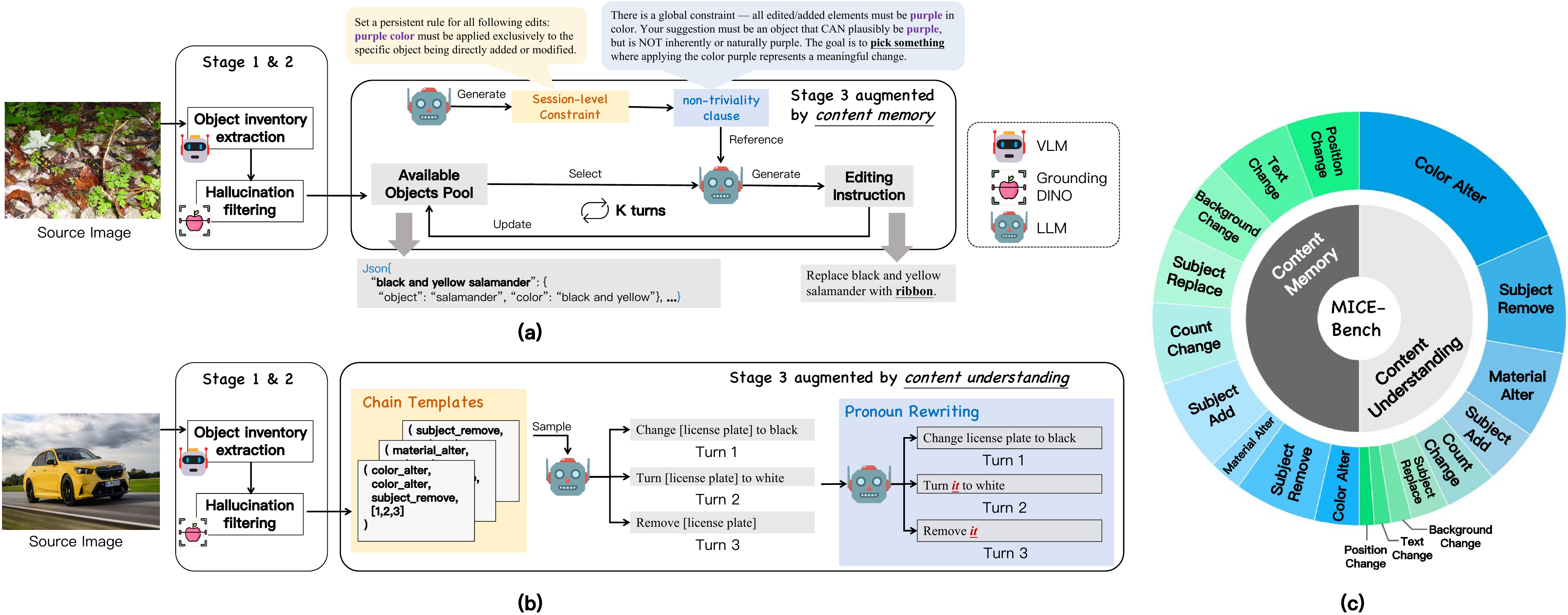} 
    \caption{ Overview of MICE-Bench. (a)-(b): Data construction pipeline. (c): Task category distribution. Detailed statistics are shown in Table~\ref{tab:task_distribution}.}
    \label{fig:data}
\end{figure}

We introduce \textbf{MICE-Bench} (\underline{\textbf{M}}ulti-turn \underline{\textbf{I}}n-\underline{\textbf{C}}ontext \underline{\textbf{E}}diting Benchmark), comprising 720 multi-turn in-context editing instances, each featuring a three-turn editing session. MICE-Bench is specifically designed to evaluate model's capabilities to resolve long-horizon dependencies featuring \emph{Content Understanding} and \emph{Content Memory} tasks (demonstrated in Figure~\ref{fig:data}). 

\textbf{Base pipeline.} We build upon EdiVal-Agent~\citep{edival}, which generates multi-turn editing instructions from unlabeled images in three stages: (1) object inventory extraction via VLM, (2) hallucination filtering with GroundingDINO~\citep{groundingdino}, and (3) 
object-pool-aware instruction generation that maintains inter-turn coherence (see Appendix~\ref{app:base-pipeline}). We enhance the pipeline with two in-context tasks.

\textbf{Content memory task.} It tests whether a model can adhere to \emph{session-level constraints} applied uniformly across the editing session (e.g., ``all added objects must be \emph{red}''). We inject such constraints into the Stage~3 instruction generator via a constrained LLM prompt, with a \emph{non-triviality clause} to prevent degenerate object proposals that trivially satisfy the constraint. See Appendix~\ref{app:constraint-prompts} for details.

\textbf{Content understanding task.} It tests whether a model can resolve pronominal coreferences across turns (e.g., ``change \emph{its} color''). We apply instruction \textit{chain templates} to ensure consecutive turns share a common anchor object, then perform post-hoc pronoun rewriting via an LLM. This design ensures pronoun references are both well-grounded and naturalistic. See Appendix~\ref{app:content-understanding} for details.

\textbf{Evaluation metrics.} Beyond standard instruction-following (IF) and content-consistency (CC)~\citep{edival}, we introduce \textbf{Global-Awareness (GA)}, a new measure specifically designed to quantify the capabilities of resolving in-context challenges. GA evaluates the model's ability to respect session-level constraints and resolve cross-turn coreferences across the editing session, with dedicated evaluation criteria. Full definitions are provided in Appendix~\ref{app:eval-metric}.

\section{Method}
\label{sec:method}

\subsection{Formulation: In-context Image Editing as an MDP}
\label{sec:mdp}

We formulate in-context image editing as a finite-horizon Markov Decision Process (MDP) $\mathcal{M} = (\mathcal{S}, \mathcal{A}, R, K)$, where $K$ denotes the number of editing turns. At each turn $t \in \{0, 1, \ldots, K{-}1\}$, the state $s_t = (Y_0, T_0, Y_1, T_1, \ldots, Y_t, T_t) \in \mathcal{S}$ denotes the full interleaved editing history, where $Y_0$ is the reference image, $\{Y_i\}_{i=1}^{t}$ are previously generated images, and $\{T_i\}_{i=0}^{t}$ are corresponding instructions. 
Crucially, $s_t$ collects the entire session history, enabling the policy to reason about long-horizon dependencies.
The action $a_t \in \mathcal{A}$ at turn $t$ is the edited image $a_t \triangleq Y_{t+1} = \pi_\theta(s_t)$. At the turn level, we treat the image generation process as a single macro-action, with intra-turn optimization detailed in Section~\ref{sec:unified_training}. The transition function is deterministic: $s_{t+1} = s_t \oplus (Y_{t+1}, T_{t+1})$, concatenating the current state with the new image and the next instruction.
A scalar reward $r_t = R(s_t, a_t)$ is received after each turn, reflecting the quality of $Y_{t+1}$. We define a compound reward as follows:
\begin{equation}\label{eqn:r}
r_t = w_{\text{IF}} \cdot \text{IF}(Y_t, T_t, Y_{t+1}) + w_{\text{CC}} \cdot \text{CC}(Y_0, Y_{t+1}) + w_{\text{GA}} \cdot \text{GA}(s_t, Y_{t+1}), 
\end{equation}
where $\text{IF} \in \{0, 1\}$, $\text{CC} \in [0, 1]$, and $\text{GA} \in \{0, 1\}$ denote instruction-following, content-consistency, and global-awareness metrics, which we directly adopt as reward components for RL training. The objective is to find $\theta$ maximizing the expected cumulative reward $\max_{\theta} \mathbb{E}_{\pi_\theta}[\sum_{t=0}^{K-1} r_t]$.

\begin{figure}
    \centering
    \includegraphics[width=0.8\linewidth]{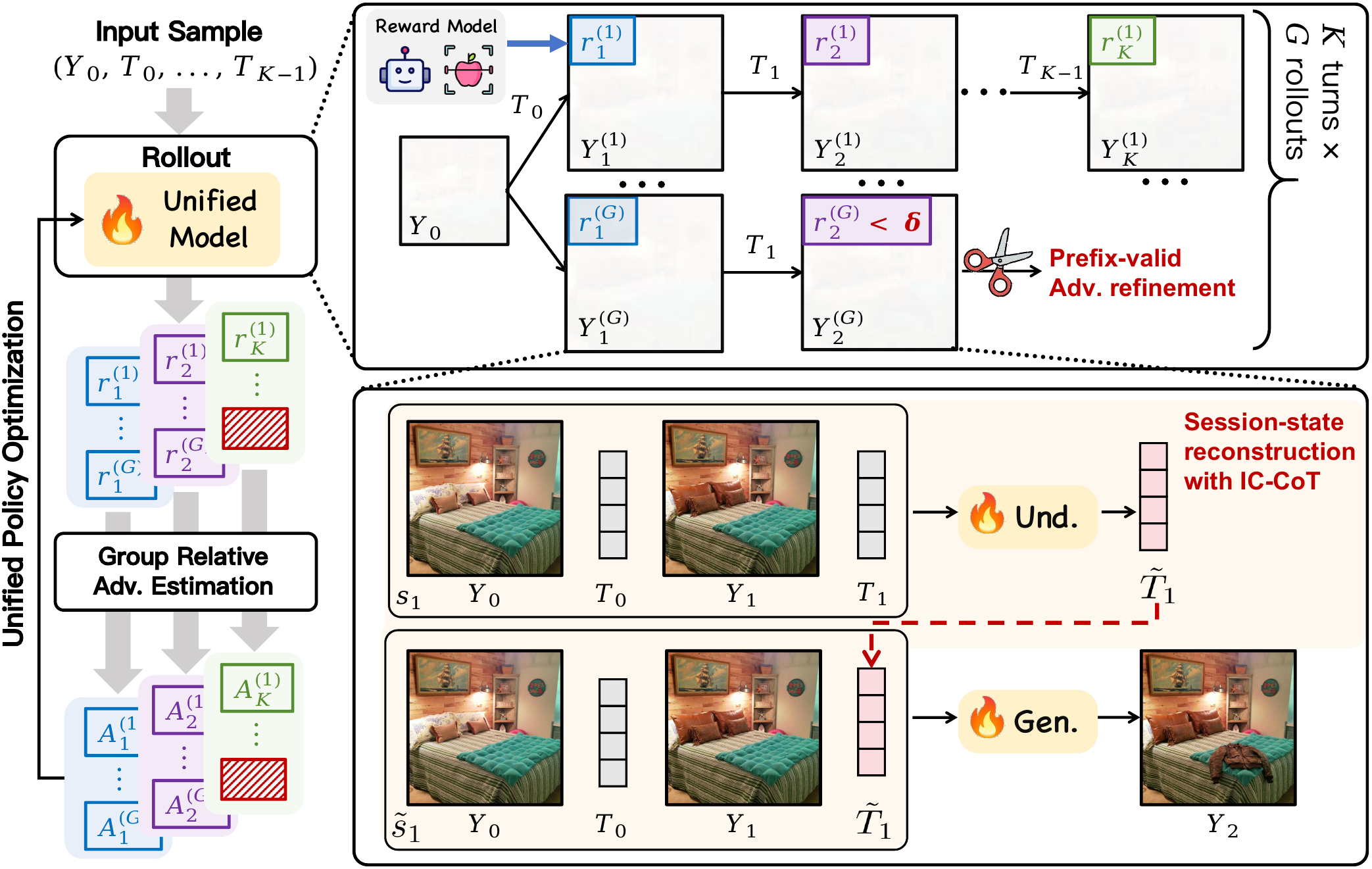}
    \caption{Overview of Edit-R2.}
    \label{fig:method}
    
\end{figure}

\subsection{Edit-R2}
\label{sec:unified_training}

Building on the aforementioned MDP formulation, Edit-R2 targets the key challenge of context loss in multi-turn image editing: the policy must act from a long interleaved history, while the decision-relevant information is a compact but latent set of active session constraints. 
Edit-R2 therefore learns to reconstruct a compact session state before generation and optimizes both the reconstruction and the visual edit with a unified trajectory-level RL objective. The overview of Edit-R2 is in Figure~\ref{fig:method}.

\paragraph{Session-state reconstruction with IC-CoT.}
Long-context attention dilution makes it progressively harder for the image generator to resolve long-horizon dependencies in the session history. Passing the full $s_t$ directly to the generator is a naive approach, but the growing token imbalance and RoPE distance decay cause attention to earlier turns to diminish, rendering significant unfaithful editing at later turns.
BAGEL's native reasoning mechanism can strengthen single-turn editing, but is fundamentally inadequate for session-aware reasoning across a historical context. Prior works of advanced CoT reasoning for visual generation~\citep{uig, uni-cot} similarly restrict to single-turn scenarios, where the reasoning scope is limited to a fixed prompt rather than an evolving session. 
We therefore introduce \textbf{IC-CoT} (In-Context Chain-of-Thought), a session-aware reasoning module that explicitly reasons over the full interleaved history $s_t$ and produces an enhanced instruction $\tilde{T}_t$ before each generation step. By distilling the relevant session context directly into the current instruction, IC-CoT compresses critical historical signals and effectively shortens the distance between the generation module and the information it needs.
At each turn $t$, the understanding expert $\pi_\theta^{\rm und}$ receives the full interleaved context $s_t$ together with a structured question $Q_{\text{CoT}}$ (see Appendix~\ref{app:method}), and produces a CoT trace followed by an enhanced instruction $\tilde{T}_t$:
\begin{equation}
\begin{aligned}
    (\tilde{T}_t,\; \mathbf{w}_t,\; \log \pi_\theta(\mathbf{w}_t)) & = \pi_\theta^{\rm und}(s_t, Q_{\text{CoT}}),
\end{aligned}
\end{equation}
where $\mathbf{w}_t$ is the CoT token sequence and their log-probabilities $\log \pi_\theta(\mathbf{w}_t)$ is retained for the subsequent policy update. The enhanced $\tilde{T}_t$ replaces the raw $T_t$, yielding the refined state $\tilde{s}_t = s_{t-1} \oplus (Y_t, \tilde{T}_t)$, which is then passed to the flow-matching generation expert:
\begin{equation}
\begin{aligned}
  (Y_{t+1},\; \mathbf{z}_t,\; \log \pi_\theta(\mathbf{z}_t)) = \pi_\theta^{\rm gen}(\tilde{s}_t),
\end{aligned}
\end{equation}
where $\mathbf{z}_t$ denotes the latent trajectory of the flow-matching process.

\begin{algorithm}[t]
    \caption{Edit-R2}
    \label{alg:bagel_flow_grpo}
    \begin{algorithmic}[1]
    \State \textbf{Input:} Reference image $Y_0$, edit instructions $\{T_t\}_{t=0}^{K-1}$, unified model $\pi_\theta$, reward function $R$.
    \State \textbf{Initialize:} $\mathcal{D} \leftarrow \emptyset; s_0 \leftarrow (Y_0, T_0)$
    \For{$t \in \{0, 1, \dots, K-1\}$}
        \State Model performs IC-CoT $(\tilde{T}_t, \mathbf{w}_t, \log \pi_\theta(\mathbf{w}_t)) = \pi_\theta^{\rm und}(s_t, Q_{\text{CoT}})$
        \State Model performs image editing $(Y_{t+1}, \mathbf{z}_t, \log \pi_\theta(\mathbf{z}_t)) = \pi_\theta^{\rm gen}(\tilde{s}_t)$
        \State State transition $s_{t+1} = \tilde{s}_t \oplus \bigl(Y_{t+1},\; T_{t+1}\bigr)$
        \State Compute reward $r_t = R(s_t, a_t)$ (Eqn.~\ref{eqn:r})
        \If{$r_t < \delta$}
            \State \textbf{break}
        \EndIf
        \State Collect samples $\tau_t = (\tilde{s}_t, \mathbf{w}_t, \log \pi_\theta(\mathbf{w}_t), \mathbf{z}_t, \log \pi_\theta(\mathbf{z}_t), r_t)$
        \State $\mathcal{D} \leftarrow \mathcal{D} \cup \{\tau_t\}$
    \EndFor
    \State Repeat the for-loop above to collect $G$ rollouts
    \State Compute group normalized advantages $\{A_t\}$ from rewards in $\mathcal{D}$ (Eqn.~\ref{eqn:adv})
    \For{each $\tau_t \in \mathcal{D}$} 
        \State Optimize $\pi_\theta$ with $\mathcal{L}_{\mathrm{unified}}(\theta)$ (Eqn.~\ref{eqn:unified}).
    \EndFor
    \State \textbf{Return:} Updated $\pi_\theta$
    \end{algorithmic}
    \end{algorithm}

\paragraph{Prefix-valid advantage refinement.}
During early training, the IC-CoT module may produce incorrect rewrites that misinterpret user intent or hallucinate non-existent content, causing the downstream generation to deviate severely from the intended edit. Continuing a rollout beyond such a failure is harmful: the resulting images carry no meaningful training signal, yet VLM-based reward models are prone to assigning false-positive scores to degenerate outputs, thereby producing noisy gradients in optimization. We therefore terminate rollout $i$ at the first turn where the reward falls below a threshold $\delta$, discarding all subsequent turns. 
Formally, we define the \emph{active set} at turn $t$ as the collection of rollouts that have not been terminated up to and including turn $t$:  
\begin{equation}
\Omega_t = \{\, i \in \{1,\ldots,G\} \mid r_{t^\prime}^{(i)} \geq \delta,\ \forall\, t^\prime \leq t \,\}.
\end{equation}                                                                                                                                                                           
The advantage estimate is then refined to normalize exclusively over active rollouts:
\begin{equation}\label{eqn:adv}
  A_t^{(i)} = \frac{r_t^{(i)} - \mathrm{mean}_{j \in \textcolor{red}{\Omega_t}}\bigl(r_t^{(j)}\bigr)}{\mathrm{std}_{j \in \textcolor{red}{\Omega_t}}\bigl(r_t^{(j)}\bigr)}, \quad i \in \Omega_t.
\end{equation}                                   
This prevents low-quality trajectories from distorting the normalization baseline, and reduces unnecessary computation by avoiding generation for rollouts that have already diverged.

\paragraph{Unified multi-turn policy optimization.}

IC-CoT and the flow-matching generator are tightly coupled, as the IC-CoT trace determines a reconstructed session state, and the generator subsequently realizes a visual edit conditioned on that state. 
However, the training signal is only observed after image generation: the quality of a CoT rewrite is only observable through the downstream image it produces, yet the generator can only benefit from IC-CoT if the rewrites are well-calibrated to its generation capabilities. 
Training them in isolation with separate objectives would break this feedback loop.
We therefore treat each turn as a joint decision of the IC-CoT token sequence and the flow trajectory for effective edits to reinforce not only the visual trajectory that produced the image, but also the preceding session reconstruction that made the edit possible.
Given this factorization, we instantiate the policy updated with Unified-GRPO~\citep{liu2026unigrpo} and extends it to the multi-turn scenario by tackling the core challenge of context loss. The pseudo-code for Edit-R2 is in Algorithm~\ref{alg:bagel_flow_grpo}. Refer to Appendix~\ref{app:method} for detailed training loss.

\section{Experiments}
\label{sec:experiments}

\subsection{Setup}

\paragraph{Evaluation benchmarks and metrics.}
We primarily evaluate on \textbf{MICE-Bench}, which measures models' ability to resolve long-horizon challenges in in-context editing, with IF, CC, and GA as evaluation metrics. We report \textit{marginal task success rate} for IF and GA: For a given turn, the proportion of previously successful prompts that continue to succeed at current turn. Results are reported in the average over 4 runs.
We also report results on EdiVal-Bench~\cite{edival} and GEdit-Bench~\cite{step1x} for generalization capability (see Appendix~\ref{app:add-exp}).

\paragraph{Baselines and training details.}
We compare against closed-source models include Nano-Banana-2~\cite{banana} and GPT-Image-1 [high]~\cite{gpt-image}. For open-source models, we consider those with native in-context editing support, including OmniGen~\cite{omnigen}, VINCIE~\cite{vincie}, BAGEL~\cite{bagel} and its variants.
We also evaluate editing-specific models Qwen-Image-Edit~\cite{qwen_image_edit} and Flux.1 Kontext~\cite{flux_kontext}. Since they do not inherently support in-context editing, we construct their input for each turn $t$ by concatenating $(Y_0, T_0, \dots, T_t)$ to approximate the in-context setting for fair comparison.
We build Edit-R2 upon BAGEL~\cite{bagel} and fine-tune with LoRA.
We use a training set comprising 10K in-context editing instances, generated via the same pipeline as MICE-Bench.
The prefix-valid advantage refinement threshold $\delta$ is set to $0.5$. Full details including hyperparameters are provided in Appendix~\ref{app:impl-details}.

\subsection{Main Results}

\begin{table}[t]
    \centering

    \caption{Results on MICE-Bench. Avg = (IF + CC + GA) / 3. Best and second-best results among open-source models are in \textbf{bold} and \underline{underlined}, respectively. Closed-source model scores that fall below Edit-R2 are marked in red. $^\dagger$Models not originally support in-context editing.}

    \label{tab:main}

    \resizebox{\textwidth}{!}{%

    \begin{tabular}{l cccc | cccc | cccc}

    \toprule

    & \multicolumn{4}{c}{\textbf{Turn 1}} & \multicolumn{4}{c}{\textbf{Turn 2}} & \multicolumn{4}{c}{\textbf{Turn 3}} \\

    \cmidrule(lr){2-5} \cmidrule(lr){6-9} \cmidrule(lr){10-13}

    \textbf{Method}

    & IF & CC & GA & Avg

    & IF & CC & GA & Avg

    & IF & CC & GA & Avg \\

    \midrule

    \multicolumn{13}{l}{\textit{Closed-source models}} \\

    GPT-Image-1 [high]~\citep{gpt-image}

    & 66.31 & \cellcolor{lightred}75.22 & 61.29 & 67.61 & 60.42 & \cellcolor{lightred}68.10 & 59.93 & \cellcolor{lightred}62.82 & 62.06 & \cellcolor{lightred}60.55 & 52.76 & \cellcolor{lightred}58.46 \\

    Nano-Banana-2~\citep{banana}

    & 62.65 & \cellcolor{lightred}87.38 & 59.97 & 70.00 & \cellcolor{lightred}49.53 & \cellcolor{lightred}82.55 & \cellcolor{lightred}46.31 & \cellcolor{lightred}59.46 & \cellcolor{lightred}57.98 & \cellcolor{lightred}76.37 & \cellcolor{lightred}52.43 & \cellcolor{lightred}62.26 \\

    \midrule

    \multicolumn{13}{l}{\textit{Open-source models}} \\

    Qwen-Image-Edit [2511]$^\dagger$~\cite{qwen_image_edit}

    & \textbf{58.62} & 81.20 & \textbf{52.84} & \underline{64.22}

    & \underline{46.47} & 77.58 & \underline{46.50} & 56.85

    & 25.04  & 72.46 & 20.76  & 39.42 \\

    Flux.1-Kontext [dev]$^\dagger$~\cite{flux_kontext}

    & 43.21 & 73.25 & 38.28 & 51.58

    & 32.56 & 71.51 & 32.63 & 45.57

    & 33.90  & 68.53 & 31.87  & 44.77 \\

    OmniGen~\cite{omnigen}

    & 44.44 & 56.84 & 38.11 & 46.46

    & 37.51 & 50.44 & 32.69 & 40.21

    & 40.97  & 47.41 & 36.76  & 41.71 \\

    VINCIE~\cite{vincie}

    & 34.77 & 71.51 & 31.67 & 45.98

    & 34.68 & 63.37 & 31.92 & 43.32

    & 43.78  & 58.02 & 26.90  & 42.90 \\

    \rowcolor{lightgray}

    BAGEL~\cite{bagel}

    & 51.38 & \textbf{89.12} & 44.72 & 61.74

    & 42.12 & \textbf{85.34} & 39.56 & 55.67

    & 50.18 & \textbf{79.90} & 43.59  & 57.89 \\

    BAGEL (w/ think)

    & 45.90 & 88.39 & 42.37 & 58.89

    & 43.66 & 83.58 & 40.03 & 55.76

    & 47.80 & 77.88 & \underline{48.14} & 57.94 \\

    BAGEL (w/ IC-CoT)

    & 51.88 & 87.80 & 46.30 & 62.00

    & 46.84 & 83.82 & 46.07 & \underline{58.91}

    & \underline{50.82} & 78.21 & 46.79  & \underline{58.61} \\

    \midrule

    \rowcolor{lightblue}

    \textbf{Edit-R2}

    & \underline{54.82} & \underline{88.62} & \underline{50.73} & \textbf{64.72}

    & \textbf{60.77} & \underline{84.33} & \textbf{58.17} & \textbf{67.76}

    & \textbf{58.63} & \underline{78.51} & \textbf{58.96} & \textbf{65.37} \\

    \bottomrule

    \end{tabular}
    }
\end{table}

We summarize key observations for results on MICE-Bench (see Table~\ref{tab:main}).
\textbf{Edit-R2 achieves the best average performance across all turns among open-source models.}
On Turn~1, Edit-R2 performs competitively among models with native in-context editing support. We note that Qwen-Image-Edit demonstrates strong single-turn instruction following capabilities, reflecting its specialization in this setting.
\textbf{On the more challenging Turn~2 and Turn~3, Edit-R2 outperforms all models on average.}
Qwen-Image-Edit and Flux.1-Kontext suffer severe performance degradation due to their inability to leverage the full session history. In contrast, Edit-R2 improves substantially over BAGEL, with IF and GA gains of approximately \textbf{(+18\%, +18\%)} at Turn~2 and \textbf{(+8\%,+15\%)} at Turn~3. These results demonstrate Edit-R2's strong performance retention on long-horizon editing tasks.
Notably, GPT-Image-1 exhibits a pronounced CC decline across turns. As qualitative analysis reveals in Section~\ref{sec:qualitative}, this reflects a tendency toward free-form generation with limited pixel fidelity.

\subsection{Ablation Study}

\begin{table}[t]
    \centering
    \caption{Ablation study of Edit-R2 on MICE-Bench. Best results are in \textbf{bold}.
    TG = Text-GRPO; PV = Prefix-valid advantage refinement.}
    \label{tab:ablation}
    \resizebox{\textwidth}{!}{%
    \begin{tabular}{l cccc | cccc | cccc}
    \toprule
    & \multicolumn{4}{c}{\textbf{Turn 1}} & \multicolumn{4}{c}{\textbf{Turn 2}} & \multicolumn{4}{c}{\textbf{Turn 3}} \\
    \cmidrule(lr){2-5} \cmidrule(lr){6-9} \cmidrule(lr){10-13}
    \textbf{Method}
    & IF & CC & GA & Avg
    & IF & CC & GA & Avg
    & IF & CC & GA & Avg \\
    \midrule
    Edit-R2 (w/o IC-CoT, TG, PV)
    & 53.99 & 88.29 & 50.06 & 64.11 & 49.32 & 82.45 & 46.28 & 59.35 & 55.20 & 72.26 & 52.01 & 59.82 \\
    Edit-R2 (w/o TG, PV)
    & 52.65 & \textbf{88.86} & 47.05 & 62.85 & 51.53 & \textbf{84.72} & 51.37 & 62.54 & 56.50 & \textbf{78.99} & 55.76 & 63.75 \\
    Edit-R2 (w/o PV)
    & 54.49 & 88.77 & 50.56 & 64.61 & 55.31 & 84.36 & 53.42 & 64.36 & 58.08 & 77.95 & 57.09 & 64.71 \\
    \rowcolor{lightblue}
    Edit-R2
    & \textbf{54.82} & 88.62 & \textbf{50.73} & \textbf{64.72} & \textbf{60.77} & 84.33 & \textbf{58.17} & \textbf{67.76} & \textbf{58.63} & 78.51 & \textbf{58.96} & \textbf{65.37} \\
    \bottomrule
    \end{tabular}%
    }
    \end{table}

\begin{figure}[t]
    \centering
    \includegraphics[width=0.9\linewidth]{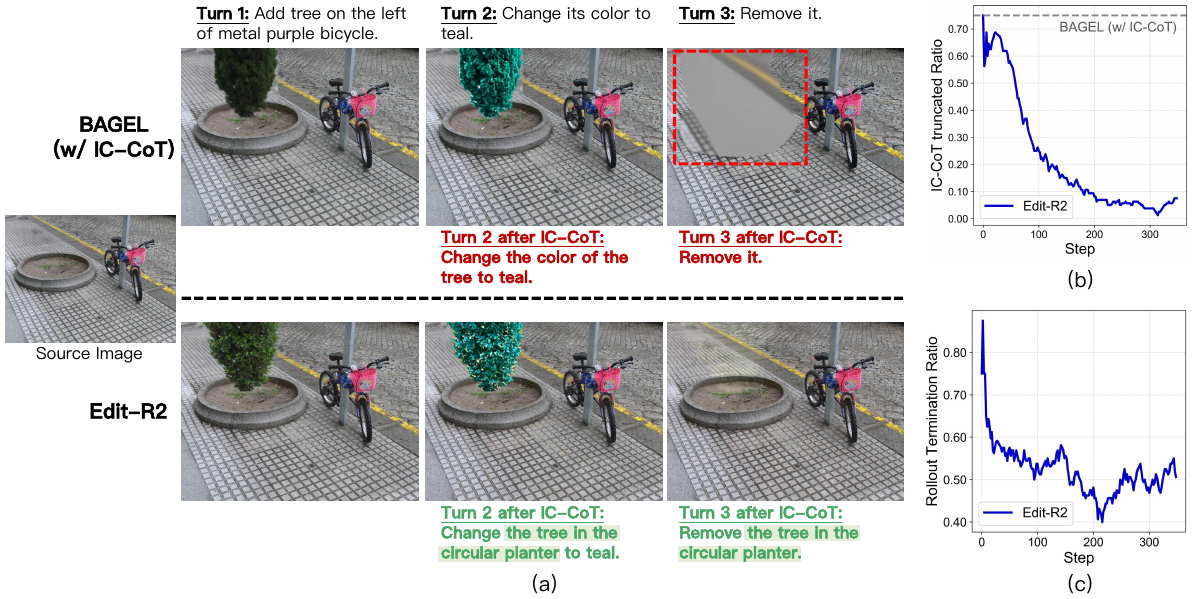}
    \caption{(a) Example of emergent IC-CoT behavior. After Edit-R2 training, IC-CoT produces more precise session-aware instruction rewrites (e.g., enriching ``the tree'' with spatial context ``in the circular planter’’). (b)-(c): Both the IC-CoT truncated ratio and rollout termination ratio decline steadily across training steps, indicating that Edit-R2 progressively stabilizes IC-CoT and reduces error propagation across turns.}
    \label{fig:cot-trend}
\end{figure}

Table~\ref{tab:ablation} reports the incremental contribution of each component in Edit-R2. The base variant without IC-CoT, TG, and PV establishes a strong foundation via Flow-GRPO alone, yet its gains are primarily concentrated at Turn~1 and diminish at later turns where the long-context challenge becomes severe. Incorporating IC-CoT yields consistent improvements at Turn~2 and Turn~3, as session-aware instruction reconstruction reduces the reasoning burden on the flow-matching module and mitigates context dilution across turns.
Further incorporating Text-GRPO---which jointly optimizes the IC-CoT module under the same reward signal---delivers additional gains at later turns, confirming that explicitly optimizing the rewriting policy amplifies its effectiveness (see Section~\ref{sec:analysis} for analysis).
Finally, PV prevents erroneous intermediate edits from corrupting the normalization baseline and propagating noise into later turns, achieving the best overall performance.

\subsection{In-depth Analysis}\label{sec:analysis}

\paragraph{Emergent instruction rewriting via Edit-R2 training.}

Following Edit-R2 training, IC-CoT exhibits emergent behavior (Figure~\ref{fig:cot-trend}(a)): it accurately identifies the target object and enriches its spatial description (e.g., ``in the circular planter''), enabling more precise editing. 
In contrast, the untrained IC-CoT generates excessive redundant reasoning, leading to output truncation and failure to complete the instruction rewriting task, resulting in poor visual quality.
As training evolves, both the IC-CoT truncated ratio and rollout-termination ratio decline steadily (Figure~\ref{fig:cot-trend}(b)(c)), confirming that Edit-R2 effectively stabilizes IC-CoT generation and validates the role of prefix-valid advantage refinement in preventing corrupted trajectories from distorting policy optimization.

\begin{figure}[!h]
    \centering
    \includegraphics[width=0.9\linewidth]{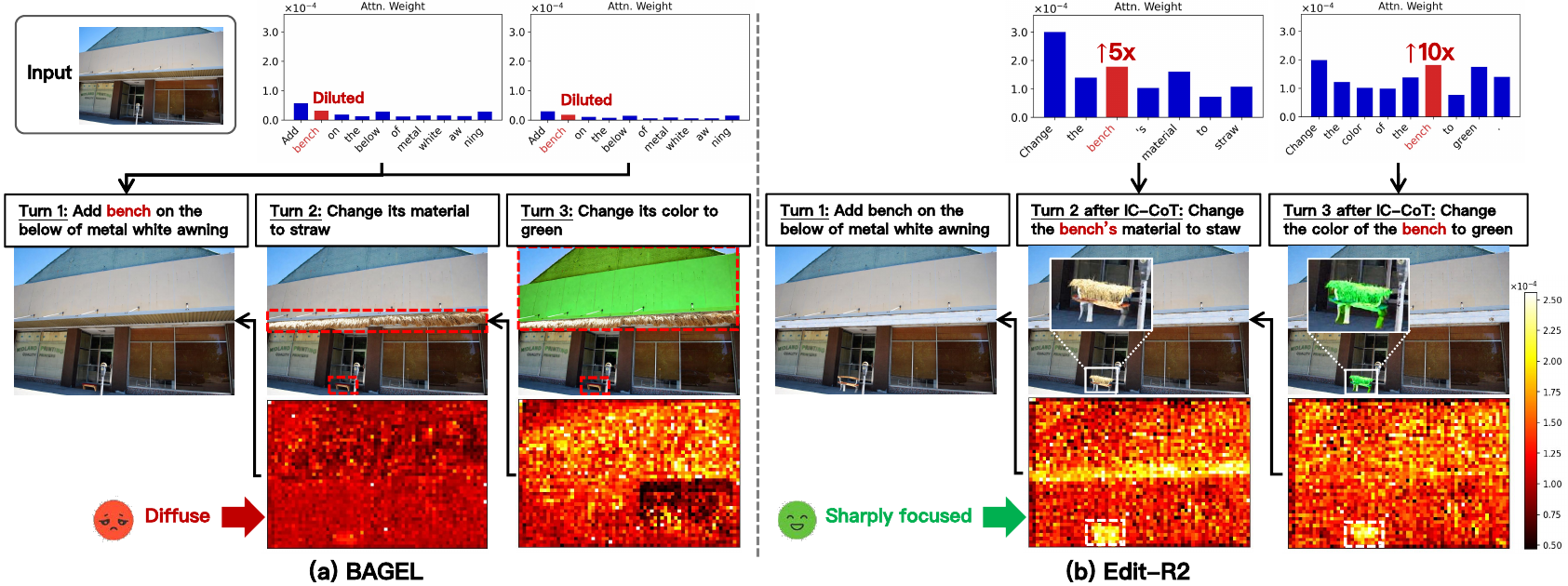}
    \caption{Attention visualization comparing BAGEL and Edit-R2. BAGEL gets lost in the long interleaved history: attention to critical tokens from earlier turns is weakened, resulting in misdirected and diffuse visual attention that leads to erroneous edits. In contrast, Edit-R2 accurately captures the operative session intent (attending $5\times$ and $10\times$ more strongly to the key object token ``bench'' at turns 2 and 3 respectively) and produces spatially focused, faithful edits.}
    \label{fig:attn}
\end{figure}

\paragraph{Attention analysis: Edit-R2 recovers session intent lost by BAGEL.} As shown in Figure~\ref{fig:attn}, we visualize text attention at the middle layer (where object-level semantic grounding is observed) and visual attention (over VAE tokens) at the last transformer layer during the velocity-prediction forward pass.
BAGEL often fails to localize the intended editing region, with misdirected and diffuse visual attention, consistent with its weak text attention on key object tokens from earlier turns where RoPE distance decay substantially weakens the signal.
In contrast, Edit-R2 exhibits sharply focused visual attention on target regions, empowered by IC-CoT which distills critical session context into the enhanced instruction.
These observations align with a text-to-visual grounding mechanism: intermediate-layer textual anchoring enriches the query with object-level semantics, which in turn guides precise spatial localization in later visual attention layers.

\subsection{Qualitative Results}\label{sec:qualitative}

\begin{figure}[tb]
    \centering
    \includegraphics[width=0.85\linewidth]{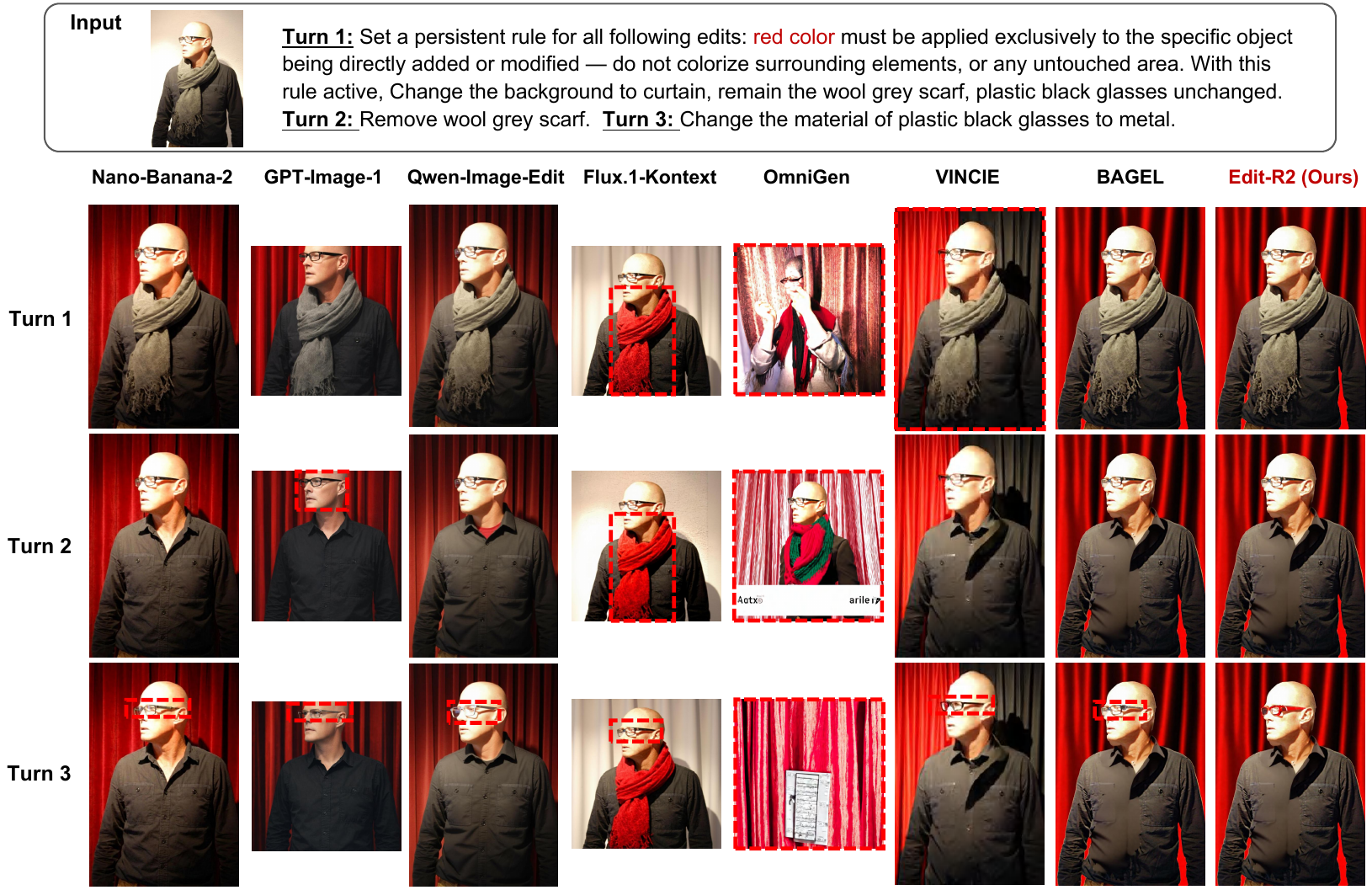}
    \caption{Qualitative comparison on \textit{Content Memory} task. Erroneous areas are in red boxes.}
    \label{fig:qualitative_cm}
\end{figure}

\begin{figure}[tb]
    \centering
    \includegraphics[width=1\linewidth]{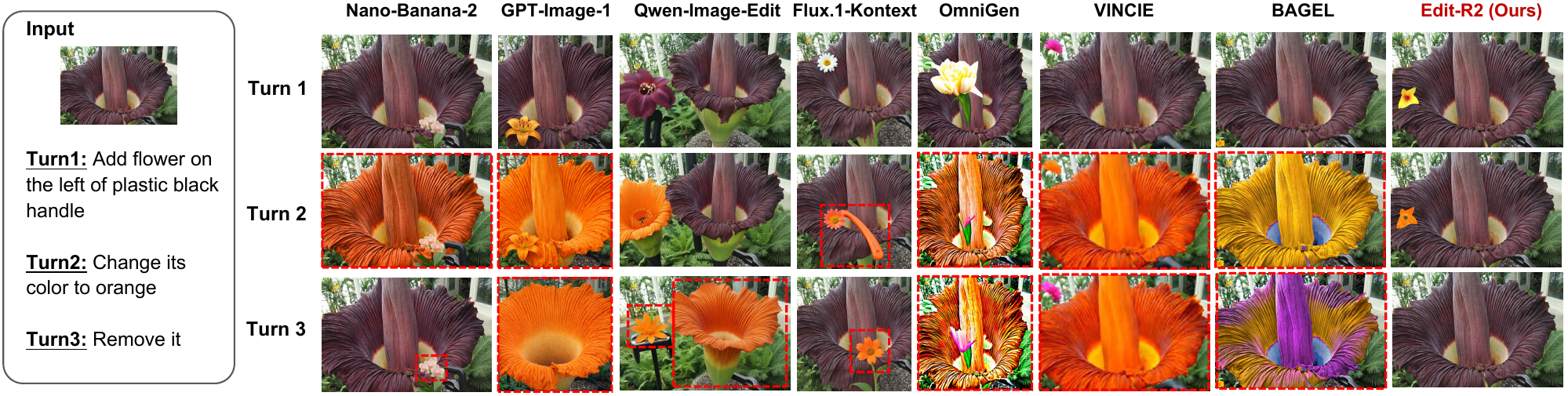}
    \caption{Qualitative comparison on \textit{Content Understanding} task. Erroneous areas are in red boxes.}
    \label{fig:qualitative_cu}
\end{figure}

Figure~\ref{fig:qualitative_cm} presents qualitative results on the content memory task. All other models completely ignore the session-level constraint ``red color must be applied'' at turn~3. In contrast, Edit-R2 successfully adheres to the constraint and changes the color of glasses to red, demonstrating improved long-horizon constraint awareness.
Figure~\ref{fig:qualitative_cu} demonstrates the content understanding task. All other models fail to correctly recolor the flower added in the previous edit to the user-specified color at turn~2, instead altering the color of unrelated regions. They also fail to correctly remove it at turn~3. Please refer to Appendix~\ref{app:add-qualitative} for more qualitative results.

\section{Conclusion}

We investigate multi-turn in-context image editing, a practical yet underexplored paradigm where the model must maintain session-level consistency across an evolving dialogue history. We propose \textbf{Edit-R2}, comprising IC-CoT for session intent reconstruction and unified multi-turn RL that jointly optimizes reasoning and image generation, alongside \textbf{MICE-Bench} for systematic evaluation. Edit-R2 yields substantial gains and achieves competitive performance against strong models.

\noindent \textbf{Limitations.} MICE-Bench covers two representative in-context editing scenarios. It is interesting for future directions to further capture the breadth of user intent in real-world interactive editing based on MICE-Bench, scale Edit-R2 to explore a broader taxonomy of in-context editing task categories.

\clearpage

%MAIN_TEXT

\bibliography{main}
\bibliographystyle{plainnat}

\clearpage
\appendix
%APPENDIX

\section{Implementation Details}
\label{app:impl-details}

\textbf{Details for experimental setup.}
During evaluation, both IF and GA employ Qwen3-VL-235B-A22B-Instruct for more aligned evaluation. Refer to Appendix~\ref{app:eval-metric} for human agreement study of judgers.
During training, we adopt a lightweight Qwen3-VL-32B-Instruct for IF and GA to accelerate online RL training. We further employ a self-ensemble strategy over the reward model to stabilize the reward signal, and apply KL regularization to prevent reward hacking. Refer to Appendix~\ref{app:add-exp} for detailed analysis.

\textbf{Hyper-parameters.} 
We fine-tune BAGEL using LoRA with rank $64$ and scaling factor $128$. LoRA adapters are applied to all attention projections and MLP layers in both the generation expert and the understanding expert, as the unified training objective (Section~\ref{sec:unified_training}) requires gradient flow through both experts. We optimize with AdamW at a learning rate of $10^{-4}$. For GRPO, we sample a group of $G{=}6$ images per prompt from a shared initial noise. Training is distributed across $4 \times 6$ NVIDIA H200 GPUs with a total batch size of 48. On each node, 2 GPUs are reserved for deploying the reward models.
We apply policy ratio clipping with threshold $\epsilon{=}10^{-5}$ and KL regularization with coefficient $\beta{=}0.01$ against the frozen reference policy.
The composite reward combines instruction following ($w_{\text{IF}}{=}0.33$), content consistency ($w_{\text{CC}}{=}0.33$), and global awareness ($w_{\text{GA}}{=}0.33$).
Each training episode spans $K=3$ editing turns.
During rollout, we use 15 denoising steps with classifier-free guidance scales of 4.0 (text) and 2.0 (image). Table~\ref{tab:hyperparams} summarizes the key hyperparameters.

\textbf{Incremental KV-Cache Update.} BAGEL naturally supports text-image interleaved history reuse via KV-Cache. During sequential editing rollout, the contextual history of preceding turns is restored from the KV-Cache at each turn. Upon completion of the current editing step, the result is incrementally updated into the KV-Cache, i.e., $s_{t+1} = \tilde{s}_t \oplus \bigl(Y_{t+1},\; T_{t+1}\bigr)$. This native reuse mechanism eliminates the need to reconstruct the full context from scratch at every turn, which constitutes a key motivation for adopting BAGEL as the base model of Edit-R2.

\begin{table}[h]
\centering
\caption{Summary of training hyperparameters.}
\label{tab:hyperparams}
\begin{tabular}{ll}
\toprule
\textbf{Hyperparameter} & \textbf{Value} \\
\midrule
\multicolumn{2}{l}{\textit{Training}} \\
LoRA rank / alpha & 64 / 128 \\
Optimizer & AdamW \\
Learning rate & $1 \times 10^{-4}$ \\
Group size & 6 \\
Total train batch size & 48 \\
Policy clip threshold ($\epsilon$) & $1 \times 10^{-5}$ \\
KL coefficient ($\beta$) & 0.01 \\
\midrule
\multicolumn{2}{l}{\textit{Reward}} \\
$w_{\text{IF}}$ / $w_{\text{CC}}$ / $w_{\text{GA}}$ & 0.33 / 0.33 / 0.33 \\
Prefix-valid advantage threshold ($\delta$) & 0.5 \\
\midrule
\multicolumn{2}{l}{\textit{Sampling \& Generation}} \\
Editing turns per episode ($K$) & 3 \\
Denoising steps (train / eval) & 15 / 50 \\
CFG scale (text / image) & 4.0 / 2.0 \\
\bottomrule
\end{tabular}
\end{table}

\section{Additional Related Work}

Comparison of MICE-Bench and existing benchmarks for image editing are summarized in Table~\ref{tab:test_set}. ``In-Context'' denotes whether the benchmark requires retention of session-level constraints across turns, beyond the immediately preceding image and instruction.

\begin{table}[tb]
\centering
\caption{Comparison of benchmarks for image editing evaluation.}
\label{tab:test_set}

\begin{tabular}{lccccc}
\toprule
\textbf{Benchmark} & \textbf{\#Size} & \textbf{\#Sub-Tasks} & \textbf{Automated} & \textbf{Multi-Turn } & \textbf{In-Context} \\
\midrule

GEdit-Bench~\citep{step1x}      & 606 & 11 & \xmark & \xmark & \xmark \\
ImgEdit-Bench~\citep{imgedit}   & 811 & 14 & \cmark & \xmark & \xmark \\
MagicBrush~\citep{magicbrush}      &  1053  & 5 & \xmark & \cmark & \xmark  \\
MSE-Bench~\citep{vincie}         &  100 & 13 & \xmark & \cmark & \xmark  \\
EdiVal-Bench~\citep{edival}          & 572 & 9  & \cmark & \cmark & \xmark \\
\midrule
\textbf{MICE-Bench (ours)} & 720 & 9  & \cmark & \cmark & \cmark \\
\bottomrule
\end{tabular}

\end{table}

\section{Additional Experimental Results}\label{app:add-exp}

\paragraph{Results on the EdiVal-Bench.} To verify the generality of our approach beyond the in-context editing setting, we further evaluate all models on the EdiVal-Bench~\cite{edival} that targets history-agnostic multi-turn editing. Results are reported in Table~\ref{tab:edival}.

\begin{table}[t]
\centering
\caption{Results on the EdiVal-Bench. IF = EdiVal-IF, CC = EdiVal-CC. Avg = per-turn average of (IF + CC) / 2.} 
\label{tab:edival}
\resizebox{0.80\textwidth}{!}{%
\begin{tabular}{l ccc | ccc | ccc}
\toprule
& \multicolumn{3}{c}{\textbf{Turn 1}} & \multicolumn{3}{c}{\textbf{Turn 2}} & \multicolumn{3}{c}{\textbf{Turn 3}} \\
\cmidrule(lr){2-4} \cmidrule(lr){5-7} \cmidrule(lr){8-10}
\textbf{Method}
& IF & CC & Avg
& IF & CC & Avg
& IF & CC & Avg \\
\midrule
BAGEL~\cite{bagel}
& 56.47 & \textbf{88.38} & 72.43 & 33.39 & \textbf{79.29} & 56.34 & 16.43 & \textbf{72.95} & 44.69 \\
\textbf{Edit-R2}
& \textbf{57.99} & 88.10 & \textbf{73.05} & \textbf{35.14} & 78.53 & \textbf{56.84} & \textbf{18.53} & 72.04 & \textbf{45.29} \\
\bottomrule
\end{tabular}%
}
\end{table}

\paragraph{Results on GEdit-Bench.} Table~\ref{tab:gedit} reports evaluation results on the GEdit-Bench~\cite{step1x} benchmark. Notably, Edit-R2 also improves on EdiVal-Bench and GEdit-Bench, indicating that the learned capabilities generalize beyond the training distribution.

\begin{table}[t]
\centering
\caption{Results on GEdit-Bench-EN (Full set). G\_SC = Semantic Consistency, G\_PQ = Perceptual Quality, G\_O = Overall Score.}
\label{tab:gedit}
\begin{tabular}{l ccc}
\toprule
\textbf{Method} & G\_SC & G\_PQ & G\_O \\
\midrule
BAGEL~\cite{bagel}
& 7.36   & 6.83   & 6.52 \\
\textbf{Edit-R2}
& \textbf{7.55}   & \textbf{6.99}   & \textbf{6.75} \\
\bottomrule
\end{tabular}
\end{table}

\begin{figure}[tb]
    \centering
    \includegraphics[width=0.9\linewidth]{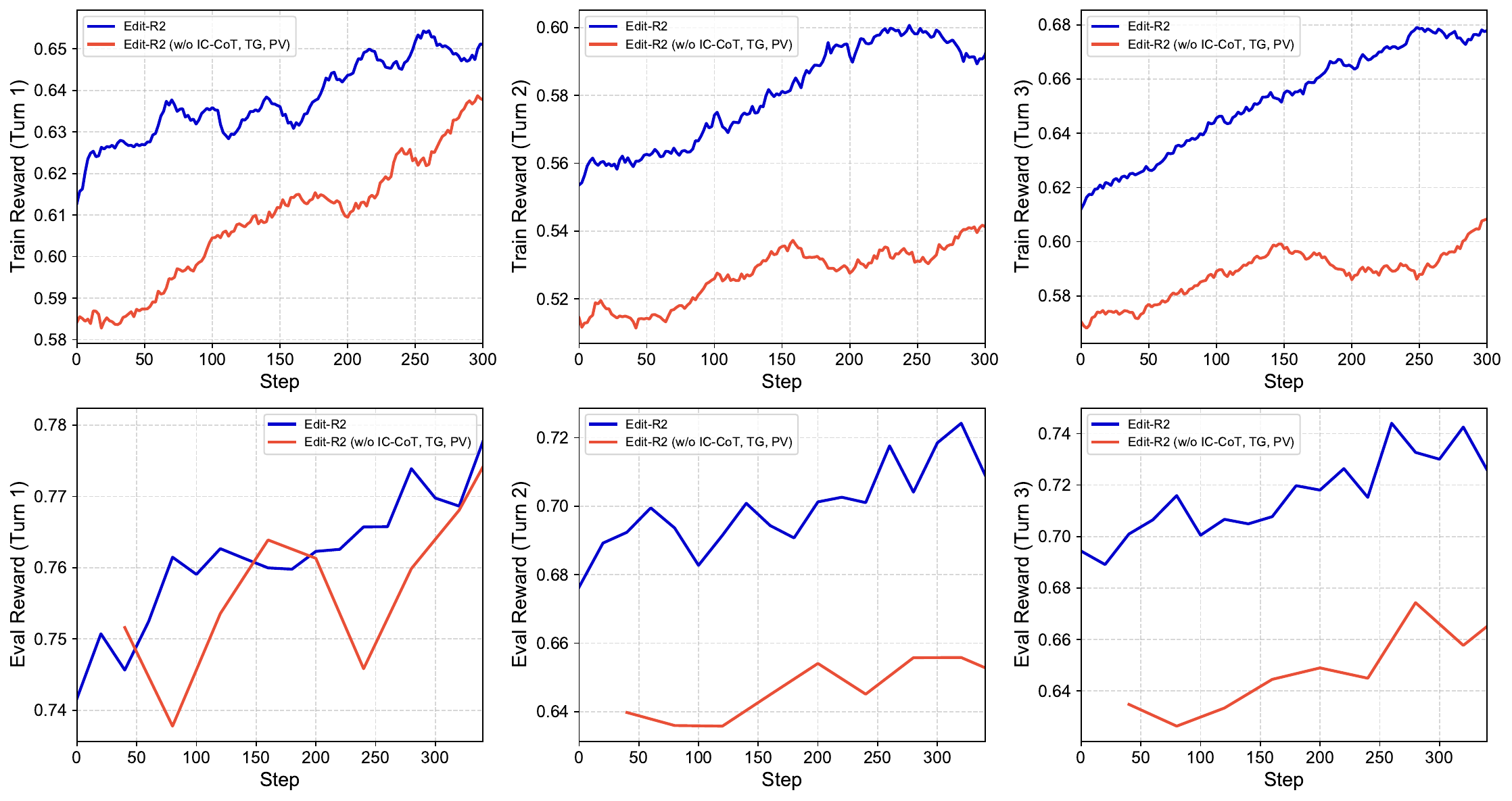}
    \caption{Training dynamics. Average rewards of IF, GA and CC are reported at each turn. Compared to the variant trained with Flow-GRPO alone, full Edit-R2 achieves more pronounced gains at turns~2 and~3, driven by IC-CoT and prefix-valid advantage refinement.}
    \label{fig:dynamic}
\end{figure}

\paragraph{Training Dynamics.} Learning curves of Edit-R2 are demonstrated in Figure~\ref{fig:dynamic}. Edit-R2 enables a steady training at each turn.

\begin{figure}[tb]
    \centering
    \includegraphics[width=0.75\linewidth]{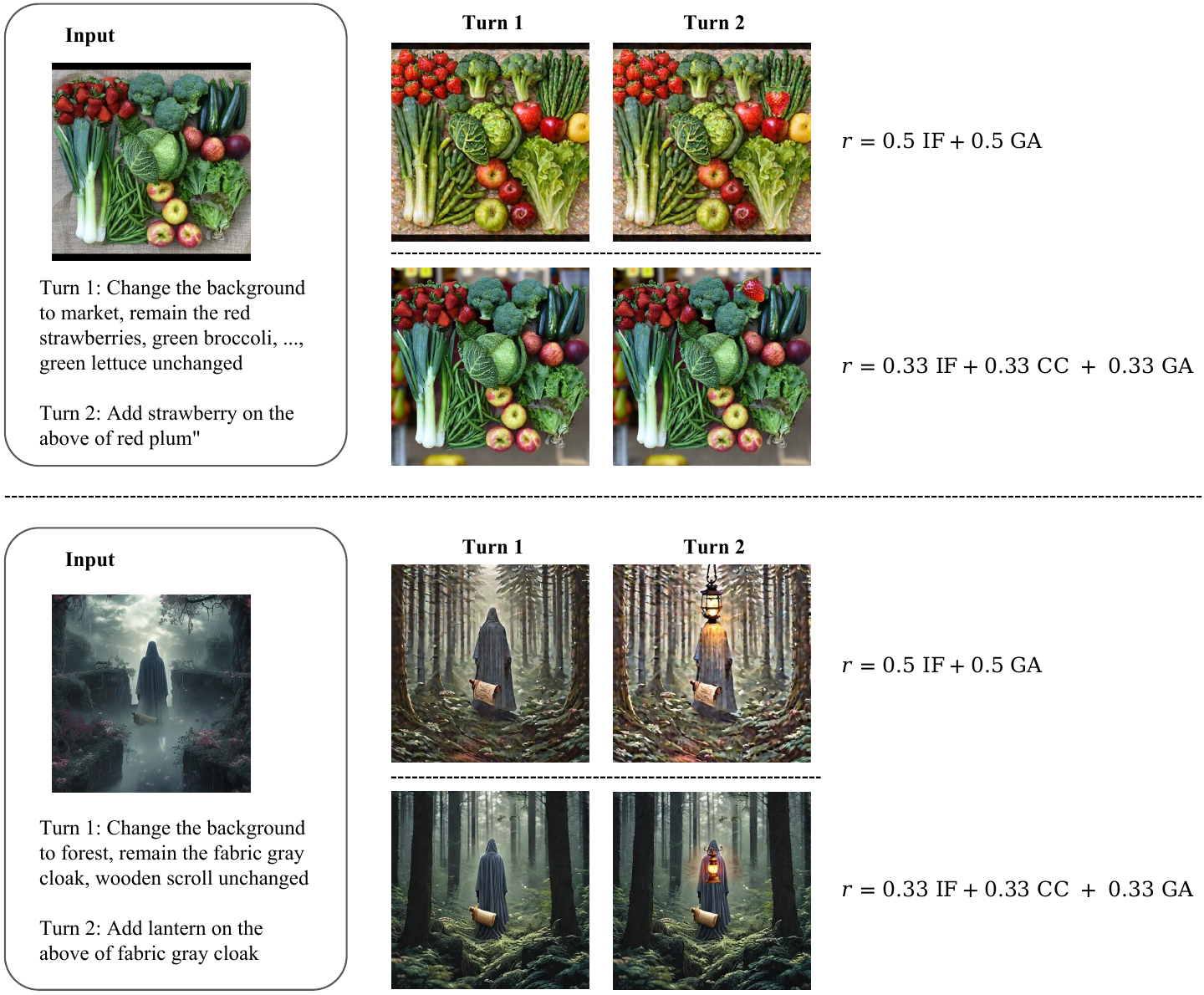}
    \caption{Ablations on reward weights. Incorporating CC in reward significantly mitigating the degradation of image quality.}
    \label{fig:cc}
\end{figure}

\paragraph{Ablations on reward weights.} Since we focus on a multiple-reward objective, determining the weight of each reward is crucial. As CC measures the similarity between the input image and the edited image, we observe that incorporating CC in the overall reward can implicitly regularize the model and prevent quality degradation. Specifically, we compare the output images generated under $r= 0.33*{\rm IF} + 0.33*{\rm CC} + 0.33*{\rm GA}$, and $r= 0.5*{\rm IF} + 0.5*{\rm GA}$. As shown in Figure~\ref{fig:cc}, CC helps prevent the generation of blurry and noise images, effectively mitigating the degradation of image quality.

\paragraph{Ablations on self-ensembled reward model.}

\begin{figure}[tb]
    \centering
    \includegraphics[width=0.75\linewidth]{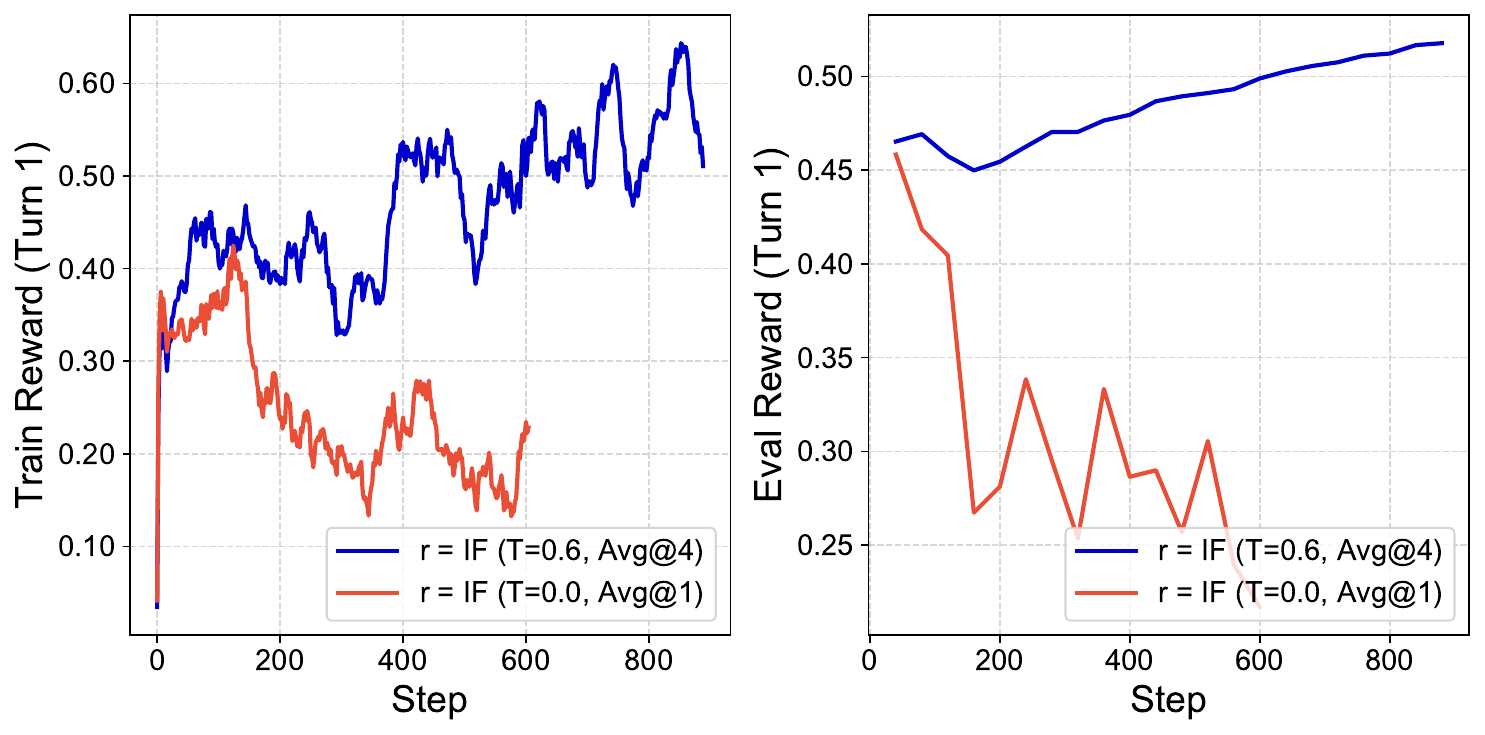}
    \caption{Ablations on self-ensembled reward model IF. $T$ denotes VLM decoding temperature; Avg@4 denotes the average score over 4 forward passes.}
    \label{fig:vote}
\end{figure}

In our initial experimental attempts, we observed that since IF and GA both rely on a VLM for scoring, the resulting reward signals inevitably carry non-trivial noise and cause training to be highly unstable, consistent with observations in prior work~\cite{editscore}.
To address this, we adopt a self-ensemble strategy: the VLM reward model replaces greedy decoding by stochastic sampling with temperature 0.6, and averages the scores from four independent forward passes to obtain the final reward signal.
Figure~\ref{fig:vote} presents training curves in the single-reward setting ($r = {\rm IF}$).
Even in this comparatively simple configuration, a greedy-sampling VLM used as IF fails to achieve effective training at turn 1: performance degrades monotonically throughout training.
In contrast, the model trained with the Avg@4 self-ensemble reward achieves steady and consistent improvement.
Consequently, all experiments in this paper adopt the Avg@4 self-ensemble strategy as the default configuration for both IF and GA.

\section{Additional Qualitative Results}\label{app:add-qualitative}

\begin{figure}[tb]
    \centering
    \includegraphics[width=0.8\linewidth]{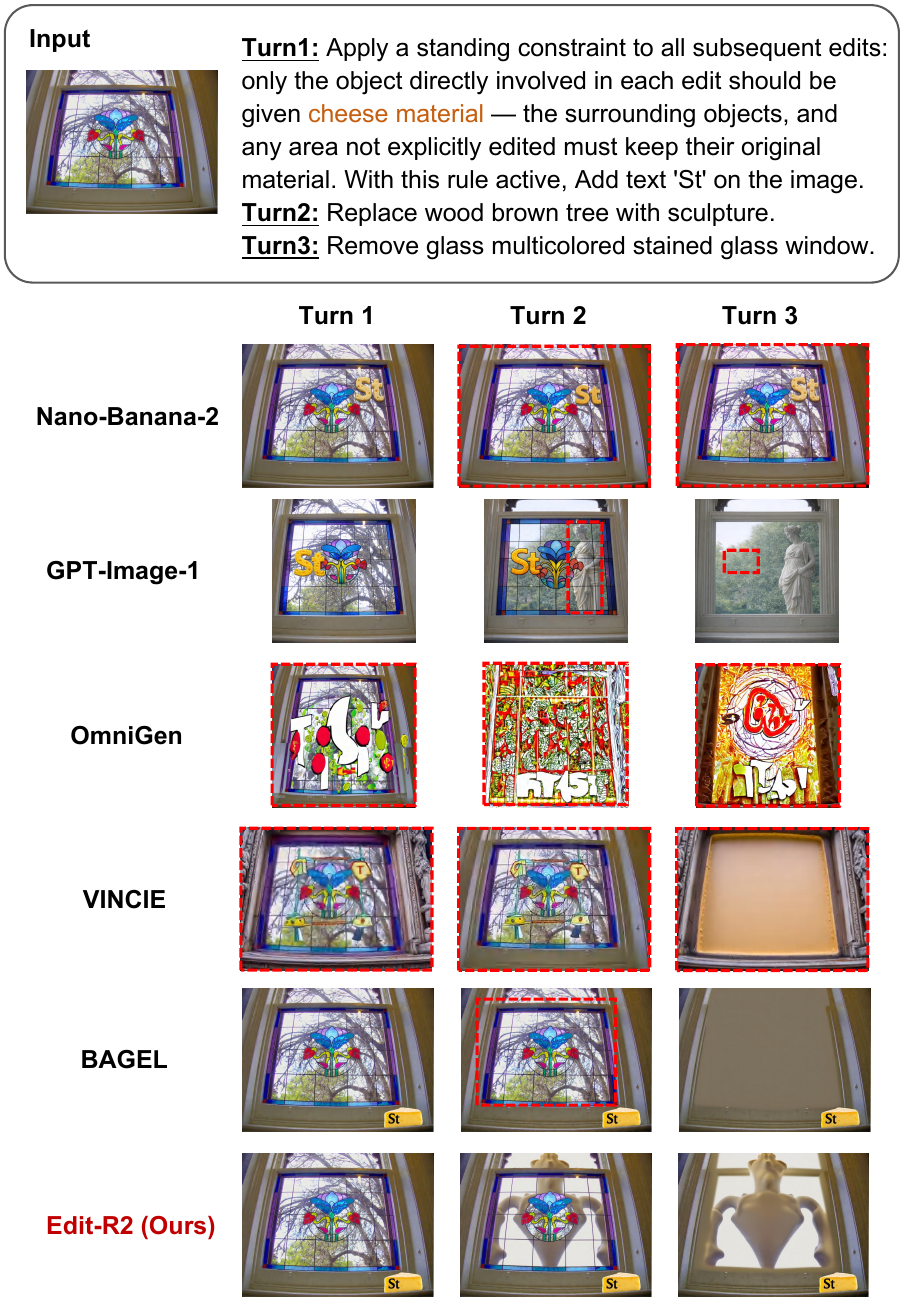}
    \caption{Additional qualitative comparison on the \textit{Content Memory} task. GPT-Image-1 and BAGEL fail to add sculpture with cheese material at turn~2; Nano-Banana-2 ignore instructions at turn~2 and turn~3.}
    \label{fig:qualitative_cm_2}
\end{figure}

\begin{figure}[tb]
    \centering
    \includegraphics[width=0.8\linewidth]{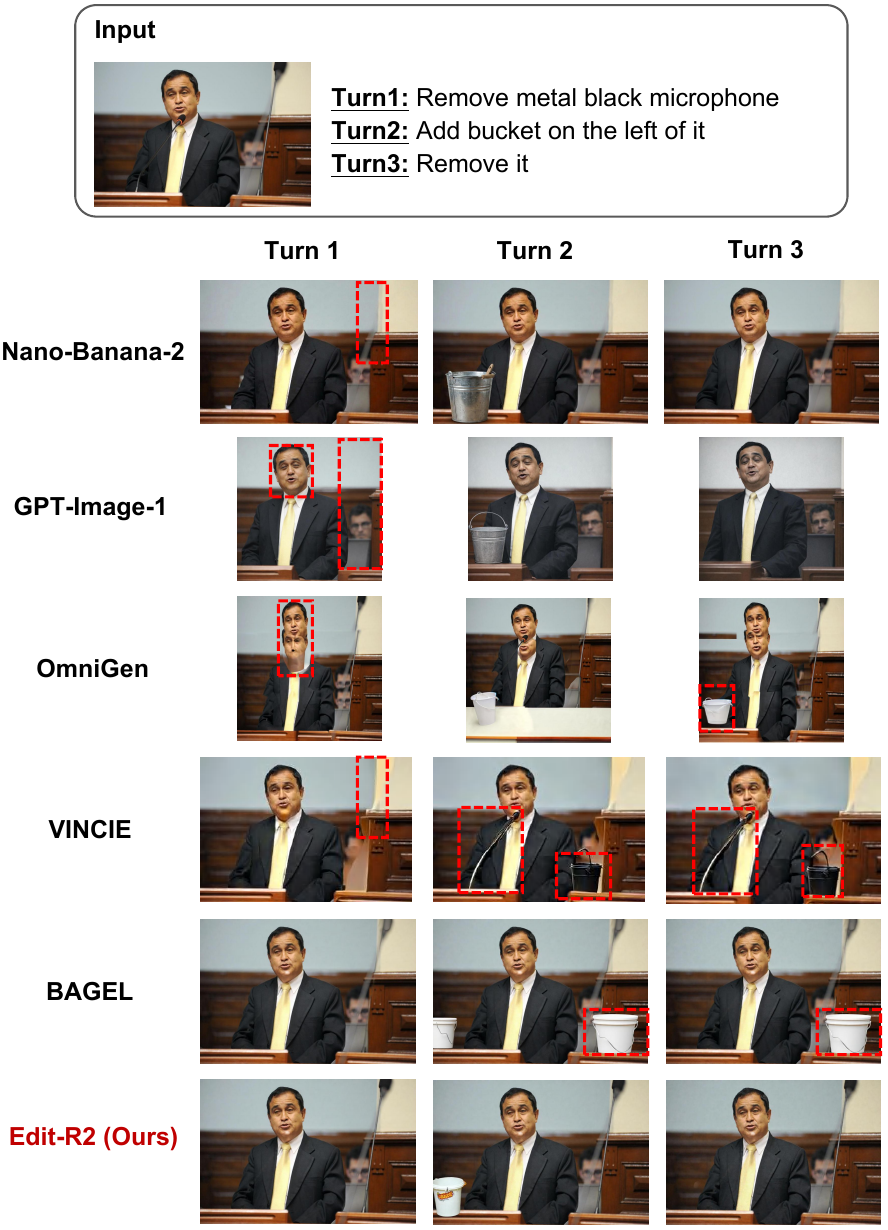}
    \caption{Additional qualitative comparison on the \textit{Content Understanding} task. All models except Nano-Banana-2 and Edit-R2 fail in this case. The two successful models correctly identify the position of the microphone removed at turn~1, place the bucket to its left at turn~2, and successfully remove it at turn~3. However, Nano-Banana-2 unintentionally removes an unrelated object at turn~1, which degrades the CC.}
    \label{fig:qualitative_cu_2}
\end{figure}

Additional qualitative results are demonstrated in Figure~\ref{fig:qualitative_cm_2} and Figure~\ref{fig:qualitative_cu_2}.

\section{Details for MICE-Bench}
\label{app:bench}

\subsection{Base Pipeline Details}
\label{app:base-pipeline}

\subsubsection{Stage 1: Object Inventory Extraction}

Given an unlabeled input image, we prompt a VLM to enumerate all clearly visible objects. For each object, the VLM describes its semantic type, dominant color, material, visible text, instance count, and foreground/background status. The output format is a structured JSON record:

\begin{itemize}
    \item \textbf{Keys} follow the template \texttt{"\{material\} \{color\} \{object\}"} (e.g., \texttt{"wooden brown table"}).
    \item \textbf{Values} encode the attribute fields listed above.
    \item A dedicated \texttt{"All Objects"} field aggregates the full inventory into a single string for downstream consumption.
\end{itemize}

This structured representation constitutes the initial \emph{object pool} $\mathcal{P}_0$ that grounds all subsequent instruction generation.

\subsubsection{Stage 2: Hallucination Filtering}

VLMs are known to hallucinate objects absent from the image~\cite{edival}. We pass each candidate object name through GroundingDINO~\cite{groundingdino}, an open-set object detector, using the object description as the text query. An object is retained only if at least one bounding box exceeds a confidence threshold. We additionally filter out detections whose bounding-box area exceeds 90\% of the image (normalized coordinates) to suppress spurious full-image matches. Each retained object is augmented with the number of detected instances and bounding-box coordinates for subsequent evaluation.

\subsubsection{Stage 3: Object-Pool-Aware Instruction Generation}

Given the verified object pool $\mathcal{P}_0$, an LLM generates a sequence of editing instructions over $K=3$ turns. At each turn, the LLM samples from a predefined taxonomy of edit types shown in Table~\ref{tab:task_distribution}.

The key mechanism is the dynamically maintained object pool $\mathcal{P}_t$, updated after each turn to reflect the expected scene state:

\begin{itemize}
    \item \textbf{Removal / Replacement.} The target object is deleted from $\mathcal{P}_t$; for replacements, the new object is inserted with specified attributes.
    \item \textbf{Addition.} The new object is inserted with default attributes.
    \item \textbf{Attribute modification.} The object entry is updated (e.g., color or material), and the object key is renamed accordingly so that subsequent instructions reference the modified entity.
\end{itemize}

By enforcing that every instruction at turn $t{+}1$ only references objects in $\mathcal{P}_t$, the pipeline guarantees that no instruction can, for example, alter the color of an object removed in a prior turn. This eliminates contradictory sequences and ensures logical coherence.

\subsection{Details for Content Memory Task}
\label{app:constraint-prompts}

\subsubsection{Session-Level Constraint Hints}

Below we reproduce the full prompt templates used by the instruction-generation LLM when a session-level global constraint is active.
Each template is injected into the corresponding edit-type prompt (e.g., \emph{subject addition}, \emph{subject replacement}, \emph{background change}).
The placeholder \texttt{\{value\}} is instantiated with the sampled constraint value at runtime (e.g., ``brown'' for color, ``fluffy'' for material).

\paragraph{Color constraint hint.}
\begin{quote}
\small\ttfamily
IMPORTANT: There is a global constraint --- all edited/added elements must be ``\{value\}'' in color.
Your suggestion must be an object that CAN plausibly be \{value\}, but is NOT inherently or naturally \{value\}.
For example, if the constraint color is ``brown'', do NOT suggest objects like ``wood'', ``log'', ``dirt'', or ``chocolate'' that are naturally brown ---
instead suggest objects like ``vase'', ``hat'', ``chair'', or ``umbrella'' that can be \{value\} but are not typically associated with that color.
The goal is to pick something where applying the color ``\{value\}'' represents a meaningful change.
\end{quote}

\paragraph{Material constraint hint.}
\begin{quote}
\small\ttfamily
IMPORTANT: There is a global constraint --- all edited/added elements must use ``\{value\}'' material.
Your suggestion must be an object that CAN plausibly be made of \{value\}, but is NOT inherently or naturally made of \{value\}.
For example, if the constraint material is ``fluffy'', do NOT suggest objects like ``teddy bear'', ``pillow'', or ``cotton candy'' that are naturally fluffy ---
instead suggest objects like ``chair'', ``lamp'', or ``vase'' that can be made fluffy but are not typically associated with that material.
The goal is to pick something where applying ``\{value\}'' material represents a meaningful transformation.
\end{quote}

\paragraph{Example: subject addition prompt with constraint hint.}
The following shows the complete prompt sent to the LLM for a \emph{subject addition} task when a color constraint ($v = \text{brown}$) is active:
\begin{quote}
\small\ttfamily
Suggest an object to add \{position\} of \{reference\_object\}.\\[4pt]
IMPORTANT: There is a global constraint --- all edited/added elements must be ``brown'' in color.
Your suggestion must be an object that CAN plausibly be brown, but is NOT inherently or naturally brown.
For example, if the constraint color is ``brown'', do NOT suggest objects like ``wood'', ``log'', ``dirt'', or ``chocolate'' that are naturally brown ---
instead suggest objects like ``vase'', ``hat'', ``chair'', or ``umbrella'' that can be brown but are not typically associated with that color.
The goal is to pick something where applying the color ``brown'' represents a meaningful change.\\[4pt]
Respond with ONLY the object name.
\end{quote}

\subsubsection{Non-Triviality Design Rationale}

A na\"ive constraint prompt (e.g., simply stating ``all objects must be brown'') causes the LLM to propose objects whose inherent nature already matches the target attribute---``log,'' ``wood,'' or ``beaver'' when brown is specified---rendering the constraint vacuous as a test of editing ability. The non-triviality clause explicitly instructs the LLM to avoid such trivial associations, ensuring the constraint genuinely challenges the editing model rather than being satisfied by the original scene content.

\subsubsection{Constraint-Aware Operation Suppression}

When a session-level constraint specifies a particular color or material, the corresponding \emph{color alter} or \emph{material alter} operations are suppressed in subsequent turns to prevent contradictory instructions (e.g., a global ``red'' constraint followed by ``change the cup to blue'').

\subsection{Details for Content Understanding Task}
\label{app:content-understanding}

\subsubsection{Instruction Chain Templates}
\label{app:chain-templates}

We curate a set of templates, each specifying a fixed sequence of three edit types and a same-object constraint indicating which turns share a common anchor object. Representative templates include:

\begin{itemize}
    \item $(\textit{color\ alter},\; \textit{color\ alter},\; \textit{subject\ remove},\; [1,2,3])$: all three turns target the same entity.
    \item $(\textit{subject\ add},\; \textit{color\ alter},\; \textit{subject\ replace},\; [1,2])$: only the first two turns share an anchor.
    \item $(\textit{material\ alter},\; \textit{position\ change},\; \textit{color\ alter},\; [2,3])$: turns~2 and~3 share an anchor.
\end{itemize}

Templates with partial overlap increase sample diversity while still guaranteeing a clear referent for pronoun substitution.

\subsubsection{Pronoun Rewriting Examples}
\label{app:pronoun-rewriting}

Given a three-turn sequence with explicit object references, an LLM rewrites turns~2 and~3 (turn~1 is unchanged, lacking preceding context). The rewriting prompt includes the original image, all three instructions, and anchor-object hints identifying the primary operand per turn. The LLM applies the following substitution rules:

\begin{itemize}
    \item \textbf{\texttt{it} / \texttt{them}}: used when the referent coincides with a prior turn's target object.
    \item \textbf{\texttt{there}}: used when referring to the spatial location of a removed entity.
    \item \textbf{\texttt{its}}: used for possessive attribute references.
\end{itemize}

Table~\ref{tab:pronoun-examples} provides representative examples for each pronoun type.

\begin{table}[h]
    \centering
    \caption{Representative pronoun rewriting examples.}
    \label{tab:pronoun-examples}
    \begin{tabular}{lp{5cm}p{5cm}}
        \toprule
        \textbf{Turn} & \textbf{Before Rewriting} & \textbf{After Rewriting} \\
        \midrule
        \multicolumn{3}{c}{\textit{Example 1: ``it'' for object coreference}} \\
        \midrule
        1 & Change the red cup to blue. & Change the red cup to blue. \\
        2 & Move the blue cup to the table. & Move \textbf{it} to the table. \\
        3 & Remove the blue cup. & Remove \textbf{it}. \\
        \midrule
        \multicolumn{3}{c}{\textit{Example 2: ``there'' for spatial reference}} \\
        \midrule
        1 & Remove the green chair in the corner. & Remove the green chair in the corner. \\
        2 & Add a lamp to the corner. & Add a lamp \textbf{there}. \\
        3 & Change the lamp to gold. & Change \textbf{its} color to gold. \\
        \midrule
        \multicolumn{3}{c}{\textit{Example 3: ``its'' for possessive reference}} \\
        \midrule
        1 & Change the material of the door to iron. & Change the material of the door to iron. \\
        2 & Change the iron door to black. & Change \textbf{its} color to black. \\
        3 & Remove the iron door. & Remove \textbf{it}. \\
        \bottomrule
    \end{tabular}
\end{table}

\subsection{Category Statistics}

Detailed task description of MICE-Bench is shown in Table~\ref{tab:task_distribution}.

\begin{table}[h]
  \centering
  \begin{tabular}{llr}
  \toprule                                                                                                          
  \textbf{In-Context Task} & \textbf{Editing Instruction Type} & \textbf{Count} \\
  \midrule                                                                      
  \multirow{9}{*}{Content Memory}                                             
   & Subject Add       & 157 \\                                                 
   & Subject Remove    & 140 \\                                                 
   & Count Change      & 138 \\
   & Text Change       & 131 \\                                                 
   & Background Change & 130 \\                                   
   & Subject Replace   & 127 \\
   & Position Change   & 125 \\                                                 
   & Color Alter       &  75 \\
   & Material Alter    &  57 \\                                                 
  \midrule                                                        
  \multirow{9}{*}{Content Understanding}
   & Color Alter       & 397 \\                                                 
   & Subject Remove    & 202 \\
   & Material Alter    & 146 \\                                                 
   & Subject Add       &  94 \\                                   
   & Count Change      &  88 \\                                                 
   & Subject Replace   &  64 \\
   & Background Change &  35 \\                                                 
   & Text Change       &  28 \\                                   
   & Position Change   &  26 \\
  \bottomrule                                                                                              
  \end{tabular}
  \caption{Task distribution of MICE-Bench. Each category (Content Memory and Content Understanding) contains 1,080 editing instructions across 360 three-turn instances, distributed among nine instruction types.} 
  \label{tab:task_distribution}                                                 
  \end{table}

\subsection{Details for Evaluation Metrics}\label{app:eval-metric}

\paragraph{IF: Instruction Following.} $\text{IF} \in \{0, 1\}$ assesses whether each edit instruction $T_t$ is successfully executed by comparing the previous-turn image $Y_{t}$ with the current edited image $Y_{t+1}$. 
Depending on the instruction type, it dispatches different verification tools. For symbolically verifiable instructions (e.g., adding, removing, or repositioning objects), GroundingDINO~\cite{groundingdino} is used to check the editing outcome against deterministic criteria. For instance, verifying a \emph{count change} instruction ``Change the number of dogs to 5'' by counting exactly five detected bounding boxes for ``dog'' in $Y_{t+1}$. For semantically verifiable instructions (e.g., changing color or background), a VLM~\cite{qwen3vl} is instead used to judge whether the intended change has been applied.

\paragraph{CC: Content Consistency.} $\text{CC} \in [0,1]$ measures the preservation of non-target content between the base image $Y_0$ and the current image $Y_{t+1}$. It consists of two components: \emph{background consistency}, which compares the background regions (excluding all object bounding boxes) across the two images, and \emph{per-object consistency}, which compares the appearance of each unchanged object individually. Both are computed using DINO-based feature similarity~\cite{dinov3}. The final CC score is the average of these two components.

\paragraph{GA: Global Awareness.} While IF and CC are sufficient for evaluating history-agnostic multi-turn editing, where each instruction depends only on the immediately preceding image, they fall short for our proposed in-context editing setting. In in-context editing, a model must maintain a session-level understanding of the user's global intent across the entire editing history, rather than treating each turn in isolation. To capture this capability, we introduce GA, it takes the full editing history $s_{t}$ and the output image of current turn $Y_{t+1}$ as input. Employing a VLM, $\text{GA} \in \{0, 1\}$ evaluates whether the \textit{content memory} and \textit{content understanding} challenges are correctly resolved.

\begin{figure}[tb]
    \centering
    \includegraphics[width=0.8\linewidth]{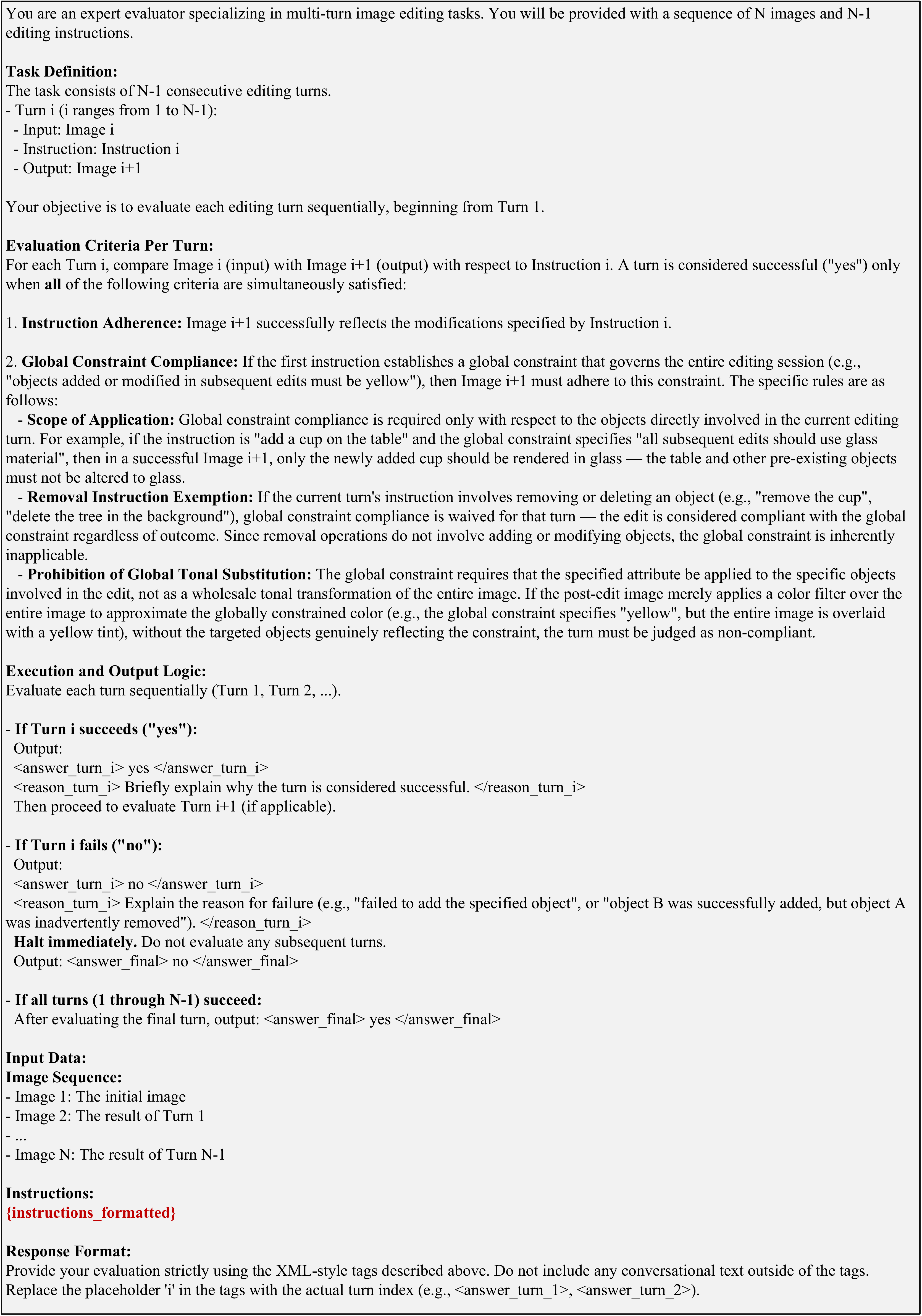}
    \caption{Prompt template used for VLM-Based GA metric on \textit{Content Memory} task.}
    \label{fig:prompt-ga-cm}
\end{figure}

\begin{figure}[tb]
    \centering
    \includegraphics[width=1\linewidth]{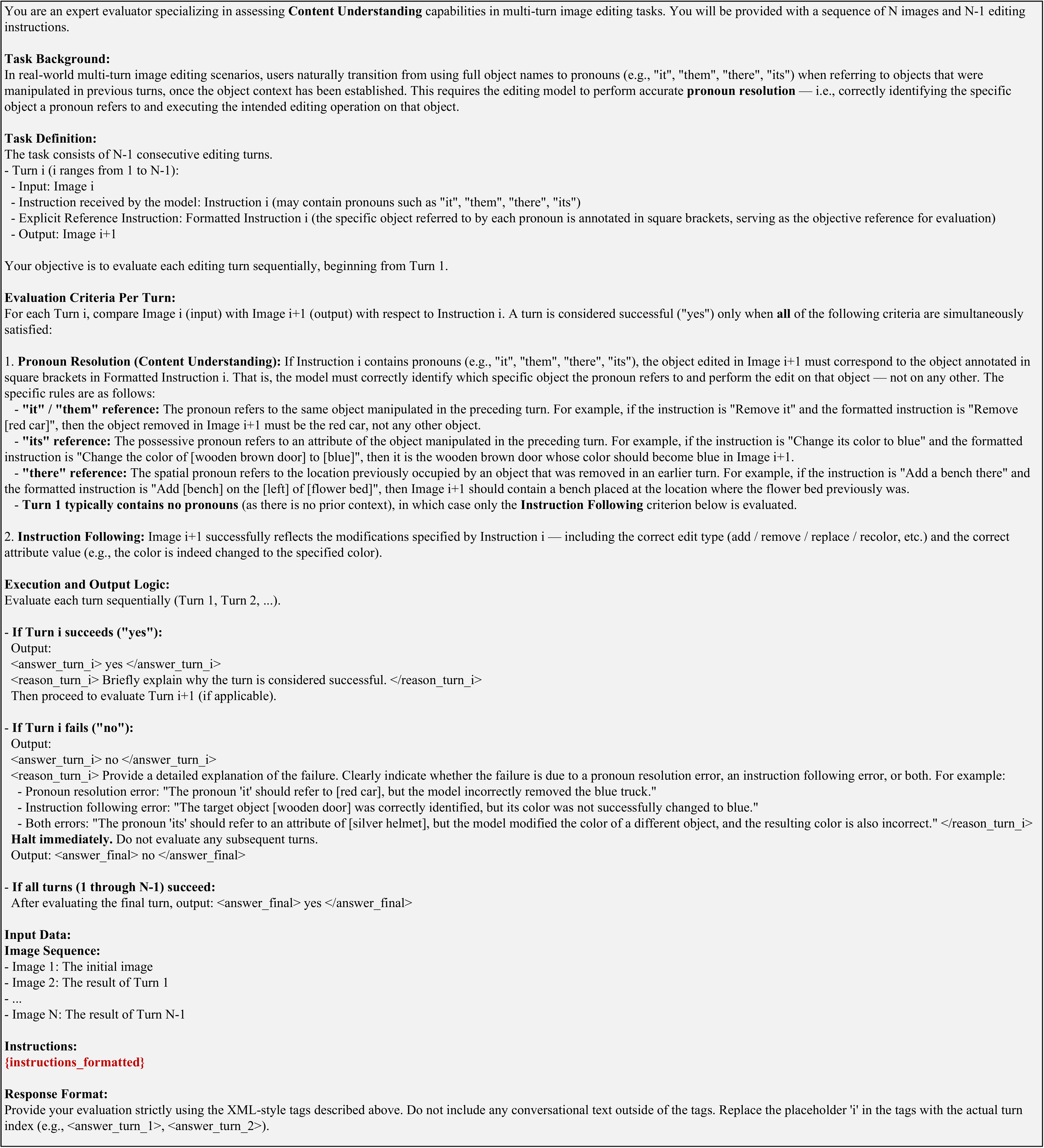}
    \caption{Prompt template used for VLM-Based GA metric on \textit{Content Understanding} task.}
    \label{fig:prompt-ga-cu}
\end{figure}

Specifically, when evaluating a multi-turn session, GA conducts a strict, turn-by-turn sequential verification. For each turn $t$, it evaluates whether the current instruction is correctly executed when transforming Image 
$Y_t$ to Image $Y_{t+1}$, while also ensuring compliance with any session-level global constraints established in earlier turns. If any single turn fails either the local instruction or the global constraint, the evaluation terminates immediately, and the overall GA score for the session is recorded as zero. This strict sequential criterion guarantees that the metric captures not only the correctness of each isolated edit, but also the model's ability to maintain global constraint awareness throughout the entire editing session.
Prompt template that used for GA on \textit{Content Memory} and \textit{Content Understanding} task are shown in Figure~\ref{fig:prompt-ga-cm} and Figure~\ref{fig:prompt-ga-cu}, respectively.

\paragraph{Human agreement study of VLM-Based GA metric.}

\begin{figure}[tb]
    \centering
    \includegraphics[width=1\linewidth]{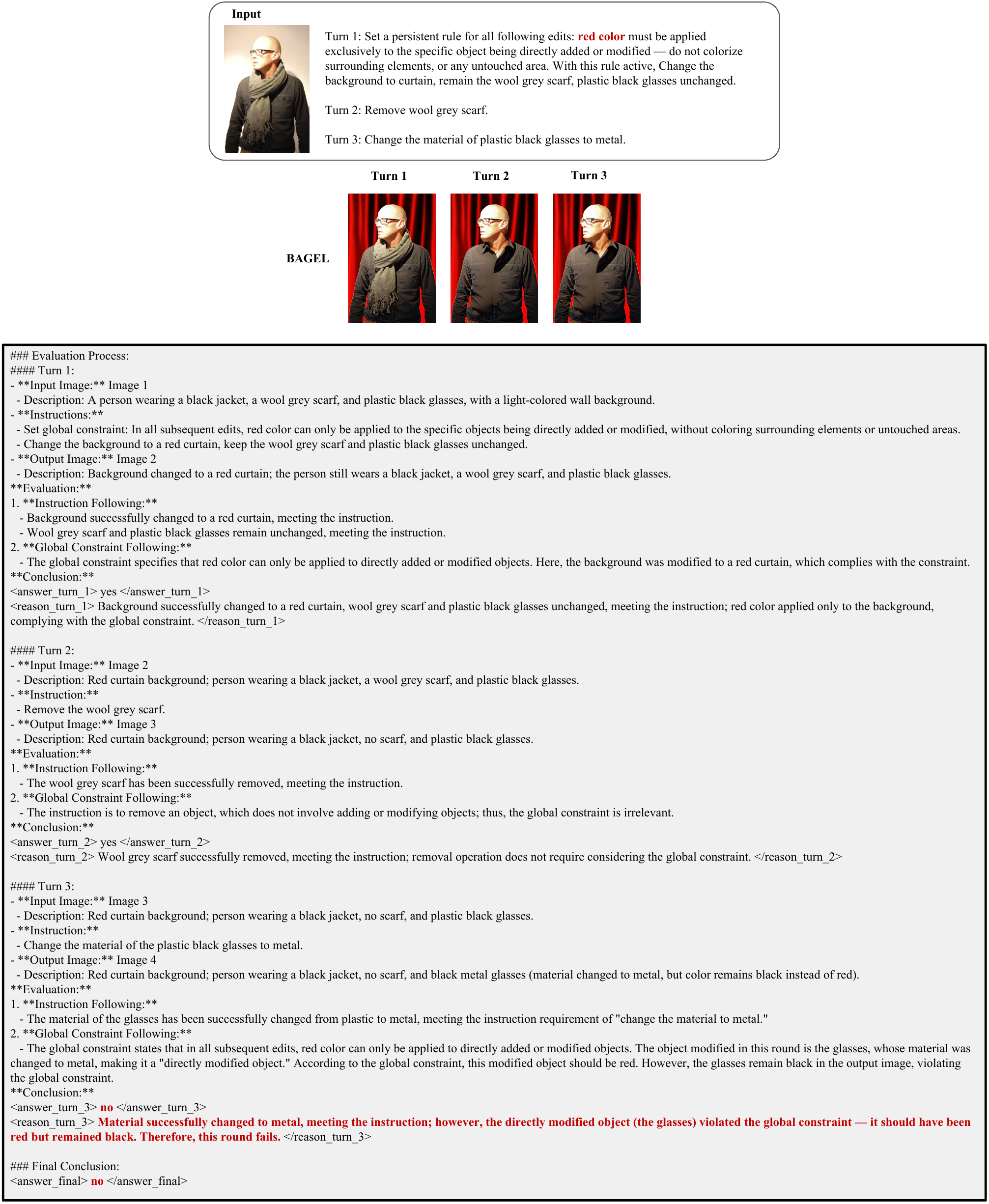}
    \caption{Example of GA output for failed editing results on content memory task.}
    \label{fig:ga-example-cm-fail}
\end{figure}

\begin{figure}[tb]
    \centering
    \includegraphics[width=0.9\linewidth]{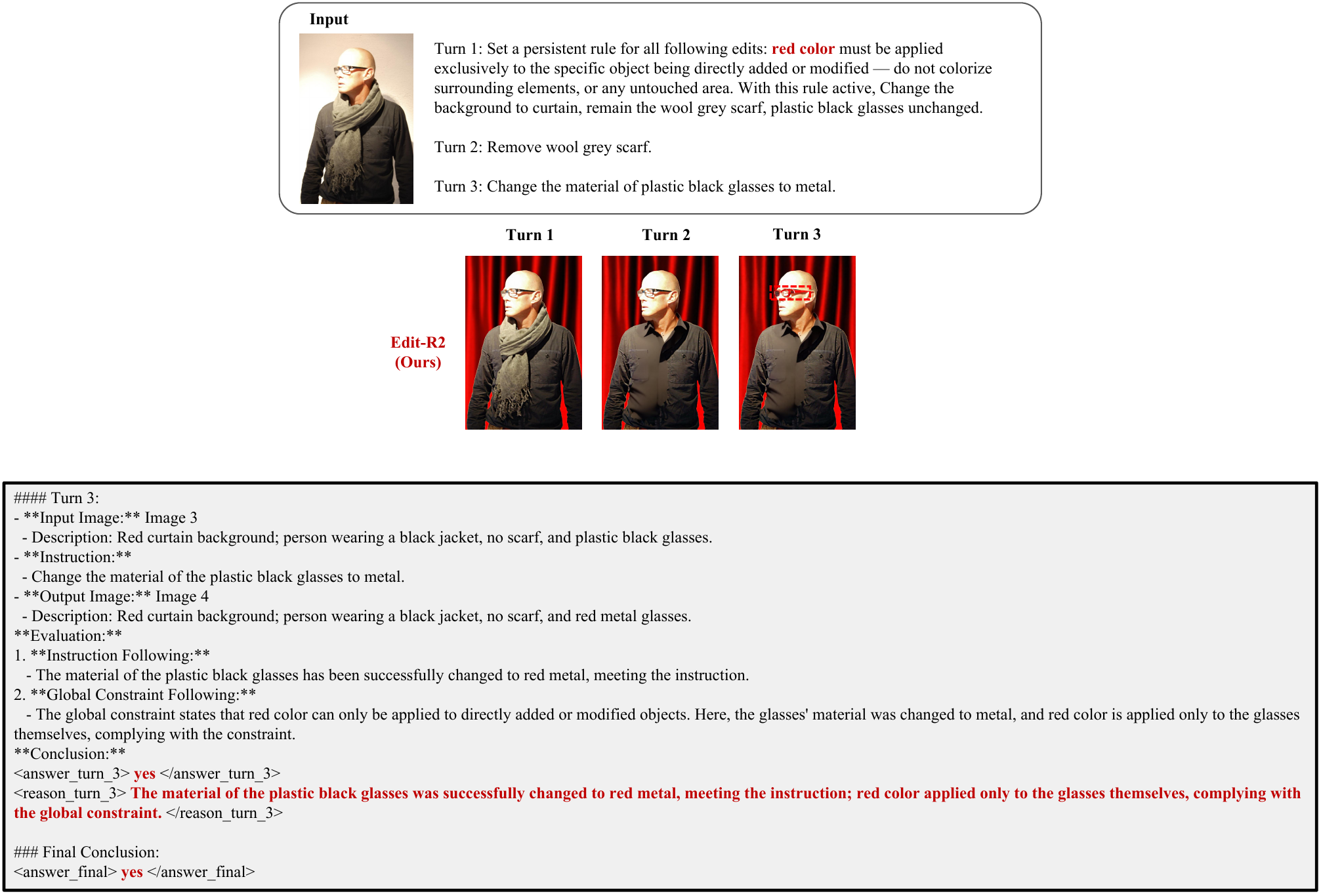}
    \caption{Example of GA output for successful editing results on content memory task. The outputs of turn 1 and turn 2 are omitted due to its similarity with Figure~\ref{fig:ga-example-cm-fail}.}
    \label{fig:ga-example-cm-succ}
\end{figure}

\begin{figure}[tb]
    \centering
    \includegraphics[width=1\linewidth]{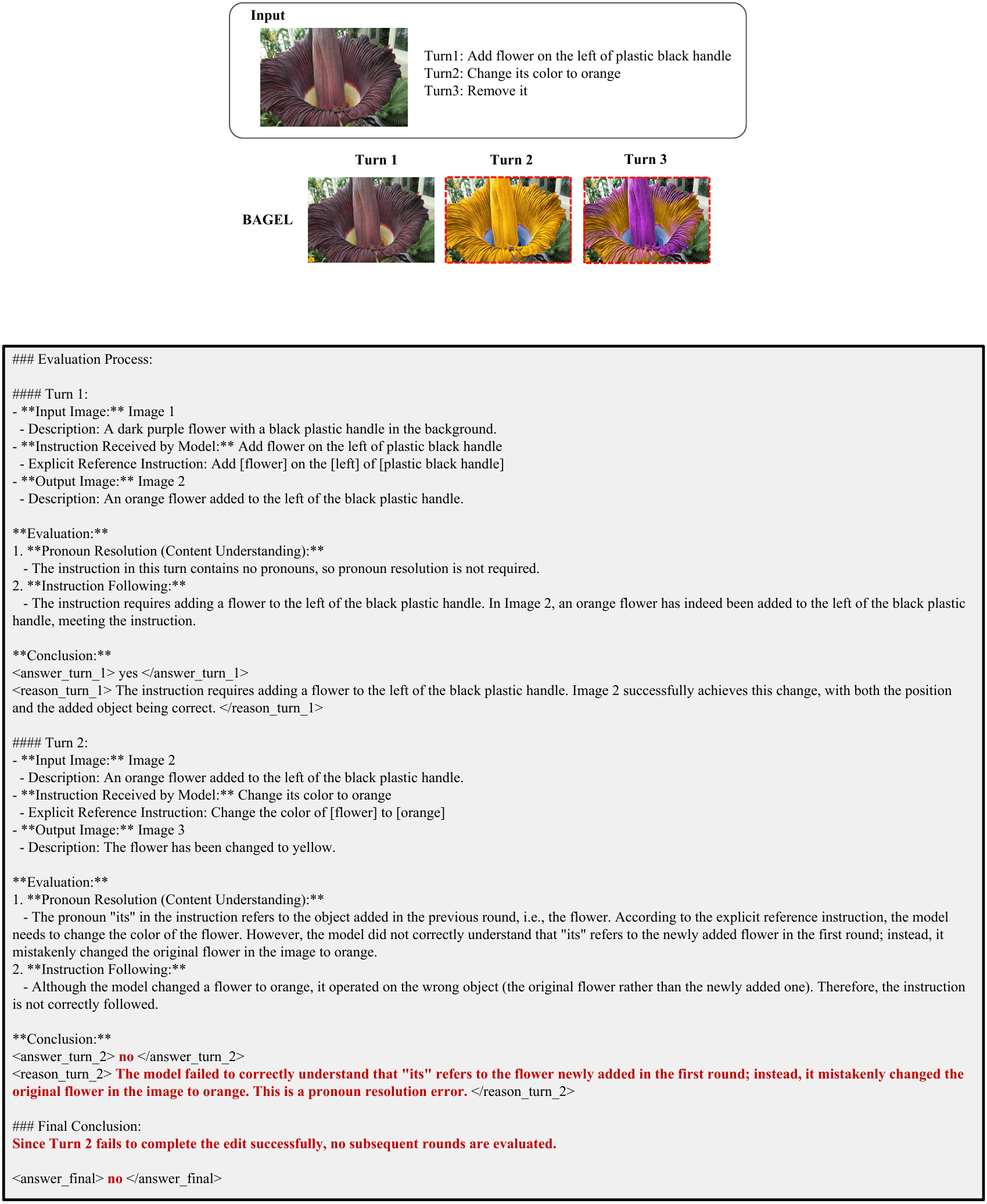}
    \caption{Example of GA output for failed editing results on content understanding task.}
    \label{fig:ga-example-cu-fail}
\end{figure}

\begin{figure}[tb]
    \centering
    \includegraphics[width=1\linewidth]{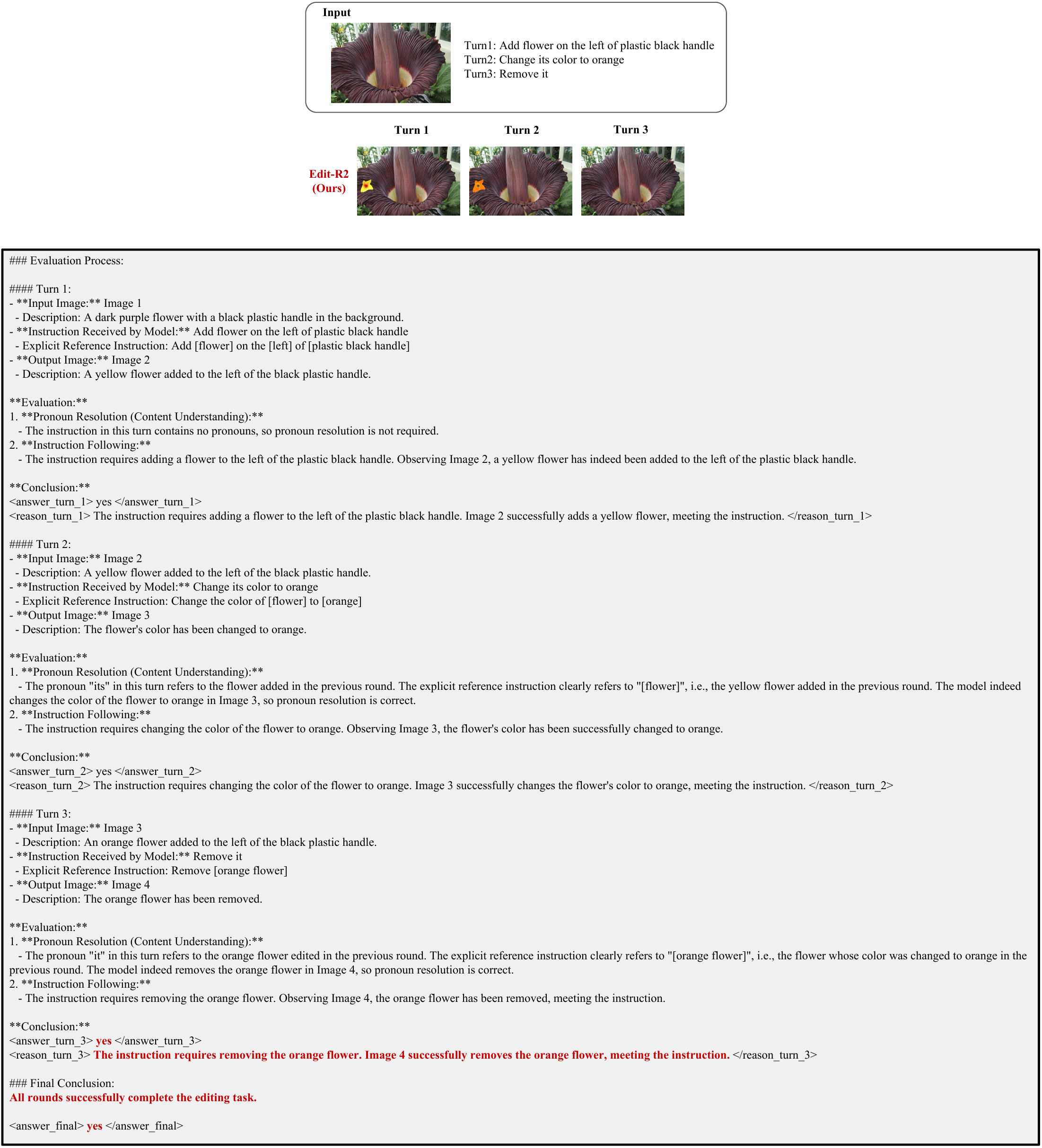}
    \caption{Example of GA output for successful editing results on content understanding task.}
    \label{fig:ga-example-cu-succ}
\end{figure}

To validate the reliability of our VLM-based GA metric, we conduct a human agreement study on 400 editing results sampled from four representative models (Nano-Banana-2, GPT-Image-1, BAGEL, and Edit-R2) covering all three turns of MICE-Bench. Human annotators were asked to judge whether each edited image satisfies the specified global constraint, without exposure to any automatic scores.
As shown in Table~\ref{tab:agreement}, GA evaluated with Qwen3-VL-235B achieves 81.5\% agreement with human judgments, substantially outperforming the 32B variant (68.3\%). The gap is most pronounced on the Content Memory subset, where the 32B judger exhibits systematic false positives---assigning GA~=~1.0 to edits that humans reject. These results confirm that the 235B judger provides a reliable proxy for human preference in evaluating global constraint compliance.
Figures~\ref{fig:ga-example-cm-fail},~\ref{fig:ga-example-cm-succ},~\ref{fig:ga-example-cu-fail}, and~\ref{fig:ga-example-cu-succ} present examples of the VLM-based GA judger's outputs on both tasks, covering both successful and failed editing results. These examples demonstrate that the GA metric aligns well with human judgments.

\begin{table}[t]
\centering
\caption{Agreement analysis of different GA evaluators with human annotations.}
\label{tab:agreement}
\begin{tabular}{lc}
\toprule
\textbf{Evaluator} & \textbf{Agreement (\%)} \\
\midrule
CLIP\_dir              & 58.9 \\
GA (Qwen3-VL-32B-Instruct)        & 68.3 \\
GA (Qwen3-VL-235B-A22B-Instruct)       & 81.5 \\
\bottomrule
\end{tabular}
\end{table}

\section{Details for Edit-R2}\label{app:method}

\begin{figure}[tb]
    \centering
    \includegraphics[width=0.75\linewidth]{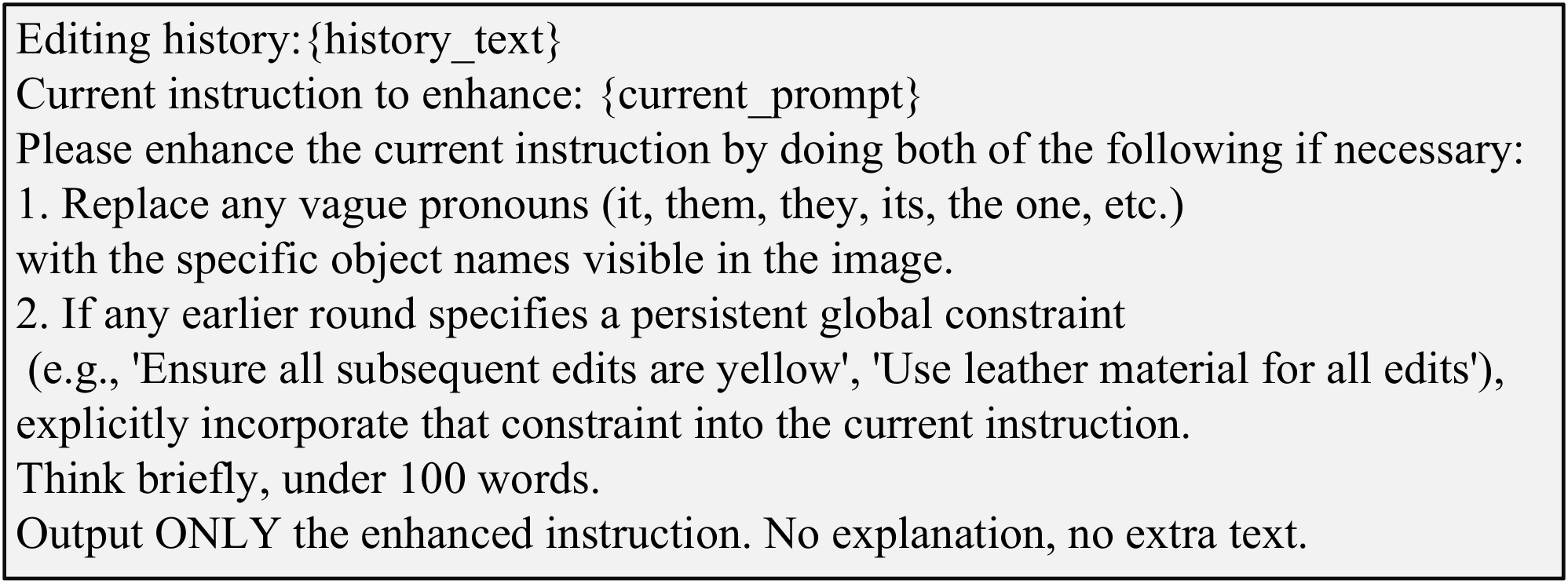}
    \caption{Illustration of $Q_{\text{CoT}}$.}
    \label{fig:q_cot}
\end{figure}

\textbf{Details for $Q_{\text{CoT}}$.} $Q_{\text{CoT}}$ guides the thinking expert to conduct two types of operations:
\begin{enumerate}
    \item \textbf{Coreference resolution}: replace vague pronouns (``it,'' ``them,'' ``its,'' etc.) with the specific object names visible in the current image $Y_t$.
    \item \textbf{Global constraint injection}: if any earlier turn specifies a persistent constraint (e.g., a color or material requirement for all subsequent edits), explicitly incorporate that constraint into the current instruction.
\end{enumerate}
Examples of $Q_{\text{CoT}}$ are shown in Figure~\ref{fig:q_cot}. Note that these questions are tailored to MICE-Bench, which differ from BAGEL's original thinking system prompt (used in``BAGEL w/ think'' in Table~\ref{tab:main}).

\textbf{Unified optimization objective in Edit-R2.}
The policy is updated by minimizing a unified objective~\citep{liu2026unigrpo} that combines Flow-GRPO and Text-GRPO losses with a shared advantage:
\begin{equation}\label{eqn:unified}
    \mathcal{L}_{\text{unified}} = \mathcal{L}_{\text{Flow-GRPO}}(\theta; \mathbf{z}_t, A_t) + \lambda \, \mathcal{L}_{\text{Text-GRPO}}(\theta; \mathbf{w}_t, A_t),
\end{equation}
where
\begin{equation}\label{eqn:flow-grpo}
\mathcal{L}_{\text{Flow-GRPO}}(\theta) = -\frac{1}{G} \sum_{i=1}^{G} \frac{1}{K} \sum_{t=0}^{K-1} \min\Big( \rho_t^{(i)}(\theta) \, A_t^{(i)},
 \mathrm{clip}\big(\rho_t^{(i)}(\theta), 1{-}\epsilon, 1{+}\epsilon\big) \, A_t^{(i)} \Big),
\end{equation}
and
\begin{equation}
\begin{split}
\mathcal{L}_{\text{Text-GRPO}}(\theta) = -\frac{1}{G} \sum_{i=1}^{G} \frac{1}{K} \sum_{t=0}^{K-1} \frac{1}{|\mathbf{w}_t^{(i)}|} \sum_{j=1}^{|\mathbf{w}_t^{(i)}|} \min\Big( & \hat{\rho}_{t,j}^{(i)}(\theta) \, A_t^{(i)}, \\
& \mathrm{clip}\big(\hat{\rho}_{t,j}^{(i)}(\theta), 1{-}\epsilon, 1{+}\epsilon\big) \, A_t^{(i)} \Big).
\end{split}
\end{equation}
Here $\rho_t^{(i)}(\theta) = \pi_\theta(a_t^{(i)} \mid \tilde{s}_t^{(i)}) / \pi_{\theta_{\text{old}}}(a_t^{(i)} \mid \tilde{s}_t^{(i)})$ and $\hat{\rho}_{t,j}^{(i)}(\theta) = \pi_\theta(w_{t,j}^{(i)} \mid w_{t,<j}^{(i)}, s_t^{(i)}) / \pi_{\theta_{\text{old}}}(w_{t,j}^{(i)} \mid w_{t,<j}^{(i)}, s_t^{(i)})$ are the per-step and per-token importance sampling ratios, $\epsilon$ is the clipping threshold, and $\lambda$ controls the balance between the two losses. A KL divergence penalty $\beta \, D_{\text{KL}}(\pi_\theta \| \pi_{\text{ref}})$ can be optionally added to prevent excessive deviation from the reference policy.

\section{Broader Impacts}
\label{app:broader-impacts}

Edit-R2 advances multi-turn in-context image editing by enabling models to follow evolving user instructions while maintaining session-level consistency. On the positive side, this technology can substantially lower the barrier to expressive visual content creation, benefiting designers, educators, and artists who rely on iterative, natural-language-driven workflows.

At the same time, improvements in controllable image generation carry inherent risks. More capable editing models can be misused to produce manipulated imagery for disinformation campaigns, non-consensual image alteration, or other forms of visual deception. We acknowledge these risks and intend to release model weights and benchmark data alongside usage guidelines and, where appropriate, content filters consistent with those of the base model BAGEL. We also encourage the community to develop robust detection tools and regulatory frameworks that keep pace with advances in generative image editing.

\end{document}